%% file: article.tex
\documentclass[10pt,a4paper]{article}
\usepackage{authblk}

\usepackage[T1]{fontenc}
\usepackage[utf8]{inputenc}
\usepackage{lmodern}
\usepackage[english]{babel}

\usepackage{graphicx}
\DeclareGraphicsExtensions{.pdf}

\usepackage{url}
\usepackage{booktabs}
\usepackage{subcaption}
\usepackage[font=small,labelfont=bf]{caption}

\usepackage{amsmath}
\usepackage{amssymb}
\usepackage{units}

\usepackage{multicol}

\usepackage{randtext}
\usepackage{comment}

\usepackage[svgnames]{xcolor}

\usepackage{enumitem}
\setitemize{itemsep=-3pt}

\usepackage{units}
\usepackage{fix-cm}
\usepackage[protrusion=true,expansion=true]{microtype}

\usepackage[margin=2cm]{geometry}
\usepackage{hyperref}
\hypersetup{
	pdfborder=0 0 0,colorlinks=true,bookmarksnumbered,urlcolor=magenta,plainpages=false,
	pdfauthor={Widera et al.},
	pdftitle={Multi-classifier prediction of knee osteoarthritis progression from incomplete imbalanced longitudinal data},
	pdfkeywords={machine learning, imbalanced data, multi-class classification, clinical decision-making, patient selection, knee osteoarthritis}
}

\usepackage[noabbrev,capitalise,nameinlink]{cleveref}

\usepackage{lastpage}
\usepackage{fancyhdr}
\usepackage{ccicons}

\fancypagestyle{plain}{
	\fancyfoot[R]{}
	\fancyfoot[L]{
		\ccby \footnotesize\, Licensed under a \href{http://creativecommons.org/licenses/by/4.0/}{Creative Commons Attribution 4.0 International Licence}.
	}
}

\usepackage{titling}

\newcommand{\hr}{
	\color{DarkGoldenrod}\rule{\linewidth}{1pt}\color{Black}
}

\begin{document}

\date{}
\title{Multi-classifier prediction of knee osteoarthritis progression from incomplete imbalanced longitudinal data}
\author{
	Paweł Widera\textsuperscript{a}, Paco M.J. Welsing\textsuperscript{b},
	Christoph Ladel\textsuperscript{c}, John Loughlin\textsuperscript{d},
	Floris P.J.G. Lafeber\textsuperscript{b}, Florence Petit Dop\textsuperscript{e},
	Jonathan Larkin\textsuperscript{f}, Harrie Weinans\textsuperscript{g,h},
	Ali Mobasheri\textsuperscript{i,j,k}, Jaume Bacardit\textsuperscript{a}
}

\pretitle{
	\vspace{-5em}
	\begin{flushleft}
	\hr
	\vskip1em
	\fontsize{24}{24} \usefont{OT1}{phv}{b}{n} \color{DarkRed}
}
\posttitle{
	\end{flushleft}
}

\preauthor{
 	\begin{flushleft}
	\large \usefont{OT1}{phv}{b}{sl} \color{DarkGreen}
}
\postauthor{
	\vskip5mm
	\footnotesize\usefont{OT1}{phv}{m}{sl}\color{Black}
	\textsuperscript{a}School of Computing Science, Newcastle University, 1 Science Square, Newcastle, NE4 5TG, UK \\
	\textsuperscript{b}Department of Rheumatology \& Clinical Immunology, University Medical Center Utrecht, Heidelberglaan 100, 3584 CX, Utrecht, NL \\
	\textsuperscript{c}Merck, Frankfurter Str. 250, 64293 Darmstadt, DE \\
	\textsuperscript{d}Bioscience Institute, Newcastle University, International Centre for Life, Newcastle, NE1 3BZ, UK \\
	\textsuperscript{e}Immuno-inflammation Center of Therapeutic Innovation, Institut de Recherches Internationales Servier, Suresnes, FR \\
	\textsuperscript{f}Novel Human Genetics Research Unit, GlaxoSmithKline, Collegeville PA 19426, US \\
	\textsuperscript{g}Department of Orthopedics, University Medical Center Utrecht, Heidelberglaan 100, 3584 CX, Utrecht, NL \\
	\textsuperscript{h}Department of Biomechanical Engineering, Delft University of Technology, Mekelweg 2, 2628 CD, Delft, NL \\
	\textsuperscript{i}Department of Regenerative Medicine, State Research Institute Centre for Innovative Medicine, Santariskiu 5, 08661 Vilnius, LT \\
	\textsuperscript{j}Research Unit of Medical Imaging, Physics and Technology, University of Oulu, Aapistie 5A, FIN-90230 Oulu, FI \\
	\textsuperscript{k}Centre for Sport, Exercise and Osteoarthritis Research Versus Arthritis, Queen’s Medical Centre, Nottingham, NG7 2UH, UK \\
	\vskip6mm
	\end{flushleft}
}

\maketitle
\vspace{-6em} \hr

Conventional inclusion criteria used in osteoarthritis clinical trials are not very effective in selecting patients who would benefit from a therapy being tested.
Typically majority of selected patients show no or limited disease progression during a trial period.
As a consequence, the effect of the tested treatment cannot be observed, and the efforts and resources invested in running the trial are not rewarded.
This could be avoided, if selection criteria were more predictive of the future disease progression.

In this article, we formulated the patient selection problem as a multi-class classification task, with classes based on clinically relevant measures of progression (over a time scale typical for clinical trials).
Using data from two long-term knee osteoarthritis studies OAI and CHECK, we tested multiple algorithms and learning process configurations (including multi-classifier approaches, cost-sensitive learning, and feature selection), to identify the best performing machine learning models.
We examined the behaviour of the best models, with respect to prediction errors and the impact of used features, to confirm their clinical relevance.
We found that the model-based selection outperforms the conventional inclusion criteria, reducing by 20--25\% the number of patients who show no progression.
This result might lead to more efficient clinical trials.

\hr
\thispagestyle{plain}

\section{Introduction}

Knee osteoarthritis (OA) is a chronic degenerative joint disease characterised by cartilage loss and changes in bones underneath it, causing pain and functional disability.
The main clinical symptoms of knee OA are pain and stiffness, particularly after activity \cite{Felson2009}, leading to reduced mobility and quality of life, and eventually resulting in knee replacement surgery.
OA is one of the leading causes of global disability in people aged 65 and older, and its burden is likely to increase in the future with the ageing of the population and rise in obesity worldwide \cite{Cross2014}.

OA is a heterogeneous disease where progression spreads over several years with periods of fast changes and periods of stability \cite{Felson2012}.
A major challenge in OA drug development is effective selection of patients to the clinical trials.
In an ideal case, all selected patients would show disease progression within the trial period, and their response to the drug in trial would be properly assessed.
However, identification of patients in need of treatment, that is those with a high probability of progression, is an open problem.

To help analyse knee OA progression, the APPROACH consortium (a partnership of over 20 European clinical centres, research institutes, small enterprises and pharmaceutical companies) is running a 2-year observational study in 5 clinical centres from 4 European countries.
One of the study objectives is to discover new markers of disease progression.
The consortium recruits patients from centres with existing OA cohorts, and similarly to clinical trials, is interested in selecting only patients who will progress during the observation period.

The traditional approach to patient selection relies on expert knowledge and typically follows a set of consensus criteria defined by the American College of Rheumatology (ACR), mixed with a presence of limited joint damage (so further progression is possible) and significant pain complaints.
When these criteria are satisfied, the patient's disease is expected to progress over time.
However, the speed with which this will happen is unknown.
This is a problem for clinical trials and short-term studies, like APPROACH, in which the observation time is typically limited to about 2 years.

The main hypothesis of this article is that machine learning can be more effective at identifying progressive patients than the traditional approach.
We hypothesise that prediction models trained on historical data will be able to differentiate between patients for whom a fast progression happen during the observation period, and patients who show no progression or progress slowly and should not be selected to trials.
Throughout the course of this article we examine different algorithms and learning process configurations, to finally develop predictive models for patient selection that outperform the conventional inclusion criteria used in clinical trials.

To train the models, and verify our hypothesis, we use longitudinal data from two large studies running in parallel in Europe and North America: the Cohort Hip and Cohort Knee (CHECK) study \cite{Wesseling2016}, and the Osteoarthritis Initiative (OAI) study \cite{Eckstein2014}.
We outline a data preprocessing strategy to handle missing values and different attribute types (\cref{sec:preprocessing}), and we define four classes of patients using clinically relevant measures of OA progression (\cref{sec:classes}).
We set up the experiments that allow us to estimate the typical performance of a model on out-of-sample instances (\cref{sec:setup}) and find the best approach to handle the class imbalance present in the data (\cref{sec:imbalance}).
We choose the best performing algorithm (\cref{sec:single-model}) and test several of its multi-model / multi-label variants to further improve performance (\cref{sec:multi-model}).
We select the most effective configuration of parameters and train the final models on all data and estimate their performance (\cref{sec:parameters}).
Then we interpret the behaviour of these models, by looking at the individual features contribution to the model output, and assess their clinical relevance (\cref{sec:interpretation}).
Next, we simulate two patient selection scenarios and compare the best model results against the selection with conventional clinical classification criteria (\cref{sec:recruitment}).
Finally, we include a discussion on limitations, the experiment design choices, related literature, and future work (\cref{sec:discussion}).

\section{Materials and methods}

\subsection{Datasets}

The CHECK cohort data used in this article were contributed by the CHECK steering committee (available upon request at \url{http://check-onderzoek.nl/}).
Specifically, we used the clinical and X-ray image assessment (radiographic scoring and KIDA features \cite{Marijnissen2008}) data.

The OAI cohort data used in the preparation of this article were obtained from the Osteoarthritis Initiative (available at \url{http://www.oai.ucsf.edu/}).
Specifically, we used the clinical, X-ray image assessment (semi-quantitative readings and joint space width measurements) and outcomes (knee replacements) data.

Both of these cohorts have been studied for over 10 years, and collected longitudinal data with typically yearly updates.
For both cohorts we used the time points between the study baseline and the 8 year follow-up, for which the joint space width measurements (used in class definition, see \cref{sec:classes}) were available (see summary in \cref{tab:datasets}).
To maximise the size of the training set, instead of using only the baseline and 2 year follow-up time points, for every patient we used all available periods that were at least 2-year long (some periods were longer than two years, e.g. between CHECK time points 2--5 or 5--8).
As a consequence, each instance in the training set represented a period, not a patient.
We excluded all periods after a knee replacement, to avoid problems with a change in meaning of some attributes (e.g. pain would no longer be related to the knee but to issues with the prosthesis).

\begin{table}[h]
\centering
\begin{tabular}{lrrrrr}
\toprule
	& \textbf{patients} & \textbf{periods} & \textbf{attributes} & \textbf{used timepoints} & \textbf{missing values} \\
\midrule
	\textbf{CHECK} & 1\,002 & 3\,001 & 513 & 0,2,5,8 & 34\% \\
	\textbf{OAI} & 3\,465 & 16\,800 & 1\,536 & 0,1,2,3,4,6,8 & 59\% \\
\bottomrule
\end{tabular}
\caption{Summary of the main characteristics of the datasets used in this work.}
\label{tab:datasets}
\end{table}

As we show in \cref{tab:datasets}, both datasets contain a relatively large number of attributes and a small number of patients, together with a large proportion of missing values.
This introduces a challenge to the machine learning algorithms and we tried to improve this balance with additional preprocessing steps (see below).

\subsubsection{Preprocessing}
\label{sec:preprocessing}

We dropped all attributes with more than 50\% missing values and all periods with over 40\% missing attributes.
These thresholds are quite conservative, as we tried to retain as much data as possible.
We also dropped all attributes that did not vary across instances (i.e. had just a single non-null value), and thus were not useful in distinguishing between the classes.
Finally, we removed attributes that could be exploited by the model, such as dates, visit numbers, barcodes and patient and staff IDs.

As the CHECK cohort is one of the recruitment sources for the APPROACH consortium, we spent extra time analysing the reasons behind the missing values and fixing them were possible.
We filled forward values from the most recent time point, for attributes which values cannot change in the future (e.g. past diseases), and used a default value in place of a missing one where this was a reporting convention (e.g. for presence of rare disorders).

For both datasets, we assumed that all attributes with at most 10 different values are categorical.
For CHECK, we additionally went through the cohort variable guide and manually identified ordinal and continuous attributes.
This step was not practical for OAI, as its variable guide has almost 4000 pages.

We performed additional preprocessing during the model training.
We imputed missing values, using only the values found in the training set (to avoid information leaks from the test set).
We performed the imputation with the mode/mean value (for categorical/continuous attributes).
We briefly tried other methods (cluster centroids, a vote of nearest neighbours), but as they did not produce better results, we settled for the simplest method.

The final step after imputation was the one-hot encoding of nominal attributes.
That is, their replacement with dummy attributes, of which only one is ``hot'' at a time (set to 1, while others are zero).
We encoded all categorical attributes with more than 2 distinct values, unless they were known to be ordinal.

\subsubsection{Class definition}
\label{sec:classes}

The APPROACH consortium decided to use similar patient categorisation to the OAI-based FNIH biomarker study \cite{Eckstein2015}, but defined more broadly and bounded in the observation time to 2 years.
Patients were split into one non-progressive category (N), and three progressive categories related to pain (P), structure (S), and combined pain and structure (P+S).

To define the categories, the consortium relied on the measures of pain symptoms and structural damage at the beginning and at the end of a period.
Pain was measured using the pain subscale from the WOMAC self-report questionnaire \cite{Bellamy2002}, which includes perceived level of pain during 5 different activities: walking, using stairs, in bed, sitting or lying, and standing upright.
Structural progression was measured using radiographic readings of minimum joint space width (JSW) across both lateral and medial femorotibial compartments of the knee.

The exact definitions of the categories are given below:

\begin{itemize}
	\item \textbf{S period} --- a minimum total JSW must decrease by at least 0.3mm per year,
	\item \textbf{P period} --- patient must experience progressive or intense sustained pain (\cref{eq:pain}):
	\begin{itemize}
	\item pain increase of at least 5 WOMAC points per year ($\Delta p \geq 5$) on 0--100 scale,
	\item pain at the end of a period must be substantial ($p_e \geq 40$),
	\item for a rapid pain increase ($\Delta p \geq 10$), end pain can be lower ($p_e \geq 35$),
	\item sustained pain must be substantial at both the start and the end of a period ($p_s \geq 40 \cap p_e \geq 40$).
	\end{itemize}
\end{itemize}

\vskip-5mm
\begin{equation}
\label{eq:pain}
	\Big((\Delta p \geq 5 \cap p_e \geq 40) \cup (\Delta p \geq 10 \cap p_e \geq 35)\Big) \cup (p_s \geq 40 \cap p_e \geq 40)
\end{equation}

For each period, the most affected knee (with greater JSW narrowing) and maximum pain (if reported for both knees) were used in the calculation of progression.
When we could not measure the progression due to missing values, we excluded the period.
This way, the class definition was never based on imputed numbers.

We assigned a period to the \textbf{P+S} category when criteria for both \textbf{P} and \textbf{S} period were satisfied, and to the \textbf{N} category when none were satisfied.
We obtained imbalanced class distributions strongly skewed towards the non-progressive periods (see \cref{tab:classes}).

\begin{table}[h]
\centering
\begin{tabular}{lllll}
\toprule
	& \textbf{N} & \textbf{P} & \textbf{S} & \textbf{P+S} \\
\midrule
	\textbf{CHECK} & 63\% \hfill ({\scriptsize 1891}) & 12\% \hfill ({\scriptsize 358}) & 20\% \hfill ({\scriptsize 592}) & 5\% ({\scriptsize 160}) \\
	\textbf{OAI} & 74\% ({\scriptsize 12502}) & \hfill 6\% ({\scriptsize 953}) & 16\% ({\scriptsize 2719}) & 4\% ({\scriptsize 626}) \\
\bottomrule
\end{tabular}
\caption{Balance between the classes for each dataset. Exact number of periods per class is given in brackets.}
\label{tab:classes}
\end{table}

\subsection{Experimental setup}
\label{sec:setup}

All experiments were performed using the \texttt{scikit-learn} library \cite{scikit-learn} and its implementation of the machine learning algorithms.
In data preprocessing, analysis and generation of statistics, we used \texttt{pandas}\cite{pandas}, \texttt{NumPy}\cite{numpy} and \texttt{SciPy}\cite{scipy}.
For data visualisation, we used \texttt{seaborn}\cite{seaborn} and \texttt{Matplotlib}\cite{matplotlib}.

\subsubsection{Measure of performance}

The problem of patient selection is similar in its nature to a well-studied task of document retrieval.
In this task, the rare relevant documents are mixed with large number of unrelated ones, and the goal is to retrieve a maximum number of relevant documents with the best possible precision.
So what matters most, is the method performance on the relevant documents.
We, in a similar fashion, are trying to identify the relevant patients who best fit the goals of the study.

The performance in information retrieval is typically measured using the $F_1$ score \cite{Sasaki2007},
which is a harmonic mean of precision and recall, designed as a measure of classifier performance in presence of rare classes.
Precision is the probability that a (randomly selected) retrieved document is relevant.
Recall is the probability that a (randomly selected) relevant document has been retrieved.
In medical literature precision is known as positive predictive value and recall is equivalent to sensitivity.

$F_1$ score is an attempt at balancing conflicting goals, because increase in recall usually comes at a cost of introducing false positives, and therefore, reduces the precision.
Compared to the area under the ROC curve, popular in medical literature, the $F_1$ score represents a trade-off among true positives, false positives and false negatives, while ROC curve represents a trade-off between true positives and false positives alone.

Although $F_1$ score has been originally designed for binary classification, it can be extended to a multi-class case, by averaging the $F_1$ scores across classes.
Throughout this article we use weighted average of per class $F_1$ scores, with weights depending on the class instance frequency (to take into account the class imbalance).

See \cref{sec:discussion-measures} for more detailed arguments behind the choice of the performance measure.

\subsubsection{Cross-validation}

In all experiments we used out-of-sample estimation of the algorithm performance.
That is, we kept some of the instances hidden from the algorithm during training, and used them later as an independent test set.
Specifically, we followed the standard 10-fold stratified cross-validation (CV) protocol, in which the instances are split into 10 approximately equal-sized parts (folds) and the split preserves the overall class distribution within each fold.
Each fold is then used in turn as a test set, and the remaining 9 folds are used as a training set.
To score the method performance, rather than averaging the scores across all 10 folds, we pool the out-of-sample predictions together and use it to calculate a single score.

The cross-validation is repeated 10 times with different partitions into folds.
As some of the machine learning algorithms are not deterministic, we also repeat the model training (25 times) with different random seeds (the seeds remain constant across folds and cross-validation repeats).
We report typical performance of a configuration (algorithm + parameters), as a median score amongst the cross-validation repeats, where the score for each repeat is the median across all trained models.

\subsubsection{Initial experiments}
\label{sec:imbalance}

To test how well different machine learning algorithms can learn from the data, we initially simplified the problem to a case of balanced classification through down-sampling.
We fixed the size of the classes to 150 for CHECK and 600 for OAI, and drew 11 different random samples of 600/2400 instances.
For each sample we performed repeated cross-validation (as described in the previous section) using for each fold a fixed-size test set, and a subset of the training set of increasing size (10\%, 20\%, \dots, 100\%), to obtain a learning curve.

We tested six machine learning algorithms with the default parameters:
\begin{itemize}
	\item \textbf{logistic regression}\cite{Fan2008} (using one-vs-rest scheme),
	\item \textbf{multinomial logistic regression} using cross-entropy loss with L-BFGS solver,
	\item \textbf{k nearest neighbours} classifier (kNN\cite{Wu2008}) using KD tree (default $k = 5$),
	\item \textbf{support vector classifier} (SVC\cite{Chang2011}) using one-vs-rest scheme with \textbf{linear} kernel,
	\item \textbf{support vector classifier} using one-vs-rest scheme with the \textbf{Radial Basis Function} (RBF) kernel (default $C = 1.0, gamma = \frac{1}{num\_features}$),
	\item \textbf{random forest}\cite{Breiman2001} (with 100 trees (default in scikit-learn 0.22)).
\end{itemize}


For scale-sensitive algorithms (SVC and kNN) all attribute values in the training set were scaled to the [0, 1] range.

In these initial experiments, random forest (see \cref{sec:single-model}) was the best performing algorithm (in line with literature \cite{Fernandez-Delgado2014,Zhang2017}), and we focused our further experiments on it.

\subsubsection{Cost-sensitive learning}

Random forest can be made cost-sensitive by incorporation of class weights to penalise the misclassification of the minority classes (as the weights influence the node split criteria).
The cost-sensitive learning is an alternative to up/down sampling techniques that does not introduce artificial instances (as with up-sampling of the minority classes) and does not lose information (as with down-sampling of the majority classes).
And specifically for random forest, the algorithm creators have demonstrated that the weighted variant performs better on imbalanced data, than on up/down-sampled ones \cite{Chen2004}.

To test the difference in performance between the cost-sensitive and the balanced learning, we first performed a repeated cross-validation (as before) using a full imbalanced dataset while incrementally increasing the training set size.
Then we kept the imbalanced test sets unchanged, and down-sampled each of the imbalanced folds used to form the training set, to obtain a balanced training set that does not overlap with the imbalanced test set.
We repeated this procedure 11 times with different sampling seeds.
In the cost sensitive variant, we used weights inversely proportional to the class distribution in the full dataset.

The rationale behind this process is that regardless of the different training sets, the test sets have to remain the same in all cross-validation rounds, so that the performance scores obtained by the two strategies are truly comparable.
With experiments set up this way, we are able to examine whether a larger training set is more important to performance than the class balance.

\subsubsection{Multi-model methods}

As we are trying to solve a multi-class problem, where the class labels are a combination of two clinical criteria (see \cref{sec:classes}), we have tested multi-model and multi-label strategies to further improve the performance of random forest.
In particular, we first tested (1) a \textit{one-vs-rest scheme}, in which a combination of 4 independent models is used, each trained to discriminate one class from the rest, and (2) a \textit{multi-label classification} \cite{Tsoumakas2007}, in which a single model is trained to assign P and S labels independently (rather than to predict the class) that are later mapped to 4 classes.
Finally, we combined the two strategies to create (3) a \textit{duo classifier} that uses two independent models, each trained to predict a single label (P or S).
We implemented this classifier as a wrapper class on top of the random forest algorithm that predicts one of the 4 class labels, but at the same time, provides independent P and S probabilities for each instance.

\subsubsection{Parameter tuning}

To tune the configuration of the duo classifier we exhaustively searched the space of 84 combinations of three key random forest parameters in the following range:
\begin{itemize}
	\item \textbf{number of trees} $\in [100, 200, 400, 600, 800, 1000]$,
	\item \textbf{maximum tree depth} $\in [4, 5, 6, 7, 8, 9, 10]$,
	\item \textbf{split quality criterion} $\in [\textit{gini}, \textit{entropy}]$ --- (standing for Gini impurity and information gain).
\end{itemize}

Because we tried multiple models, cross-validated performance of the best configuration is an optimistically biased estimate of the performance of the final model trained on all data.
This ``multiple induction'' problem is conceptually equivalent to multiple hypothesis testing in statistics.
To estimate the unbiased performance of the final model, we used a recently proposed bootstrap-based BBC-CV protocol \cite{Tsamardinos2018}.
It is a computationally efficient alternative to the popular nested cross-validation procedure and provides good bias estimation for datasets with 100+ instances.

BBC-CV uses the out-of-sample predictions to (1) select a configuration with best performance on a bootstrapped sample of instances, and (2) score the performance of the selected configuration on the out-of-bootstrap instances only.
The returned performance estimate is the average out-of-bootstrap score over all bootstrap iterations.

As we repeat each cross-validation 10 times, we used the most robust variant of the protocol --- BBC-CV with repeats.
It includes in the estimate the results from all CV-repeats, which reduces the variance introduced by the random partitioning into folds.
The number of bootstraps in the protocol was set to 1000.

\newpage
\subsubsection{Recursive feature elimination}

To test if a reduced set of features can lead to better performance, we added an inner 3-fold cross-validation loop that selects the best subset of features to use in model training.
The inner loop operates on the training folds only.
It starts from a full set of features and eliminates the worst, one by one, until only one feature is left.
Then a subset of features that maximises inner cross-validation score is selected and used to train the model on the full training fold.

\subsubsection{Model interpretation}

As each tree in the random forest votes for a class label, it is possible to count how many times each of the features have contributed to the final decision and estimate the feature importance.
The problem with the feature importance determined in this way, is that it treats all splits in a tree equally, while the early, close to the root splits, tend to have the most impact.

Therefore, we decided not to use the feature importance provided by the random forest, but to examine each tree using the \texttt{TreeExplainer} class from the SHAP module \cite{Lundberg2018,Lundberg2017}.
It provides consistent and locally accurate (per prediction) estimates of feature influence on the model output.
It combines ideas from game theory (Shapley sampling values) \cite{Strumbelj2014} and local explanations (LIME method) \cite{Ribeiro2016} and goes beyond the impact magnitude, providing information on the direction of the influence (probability boost/reduce) in relation to the feature low/high values.

\subsubsection{Comparison to the conventional inclusion criteria}

To simulate conventional inclusion decisions, we used a logical conjunction of the following three criteria: (1) a combination of the ACR clinical classification criteria for knee OA \cite{Altman1986}, (2) the Kellgren \& Lawrence grade of OA severity \cite{Kohn2016,Kellgren1957} between 1 and 3 (inclusive), and (3) pain complaints resulting in at least 40 points score on the WOMAC questionnaire.
We applied the variant of ACR criteria that uses history, physical examination and radiographic findings.
It requires presence of (1) pain in the knee and (2) one of: age over 50, less than 30 minutes of morning stiffness, crepitus (crackling noises) on active motion and osteophytes.
We assumed the criteria are satisfied if one of the knees satisfy them.

To simulate selection with machine learning models we used two scenarios: \texttt{ML-L} (based on class labels) and \texttt{ML-P} (based on class probabilities).
Both scenarios were based on predictions made by the best configuration of the duo classifier, specifically the median score model from the median cross-validation repeat.

In the \texttt{ML-L} scenario, we selected all instances classified as progressive (predicted to belong to the P, S, or P+S class).
This scenario simplifies the task to a binary classification, and makes it comparable to the binary decision made using the conventional inclusion criteria.

In the \texttt{ML-P} scenario, for a more direct comparison, we selected the same number of instances as obtained with the conventional criteria.
We used the progression probabilities $p(S)$ and $p(P)$ returned by the model to three-way sort the instances (in a descending order) by $p(P)+p(S)$, $p(S)$, and $p(P)$.
Then we selected \nicefrac{1}{3} of instances from each sorted group (to obtain balanced representation), in that exact order, disregarding the duplicates.

\section{Results}

\subsection{Comparison of algorithms on balanced subsets}
\label{sec:single-model}

\begin{figure}[!th]
\begin{subfigure}{\textwidth}
	\includegraphics[width=\textwidth]{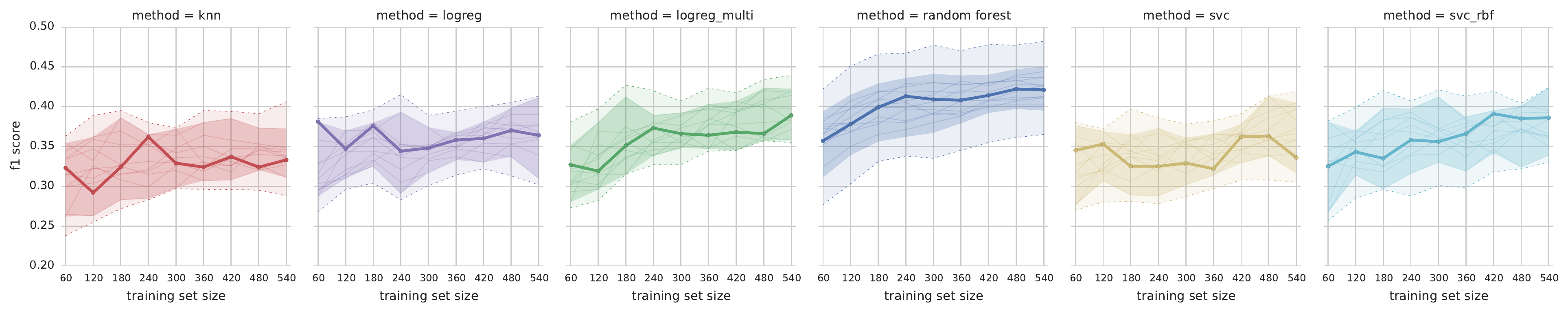}
	\caption{CHECK dataset}
	\label{fig:single-check}
\end{subfigure}
\begin{subfigure}{\textwidth}
	\includegraphics[width=\textwidth]{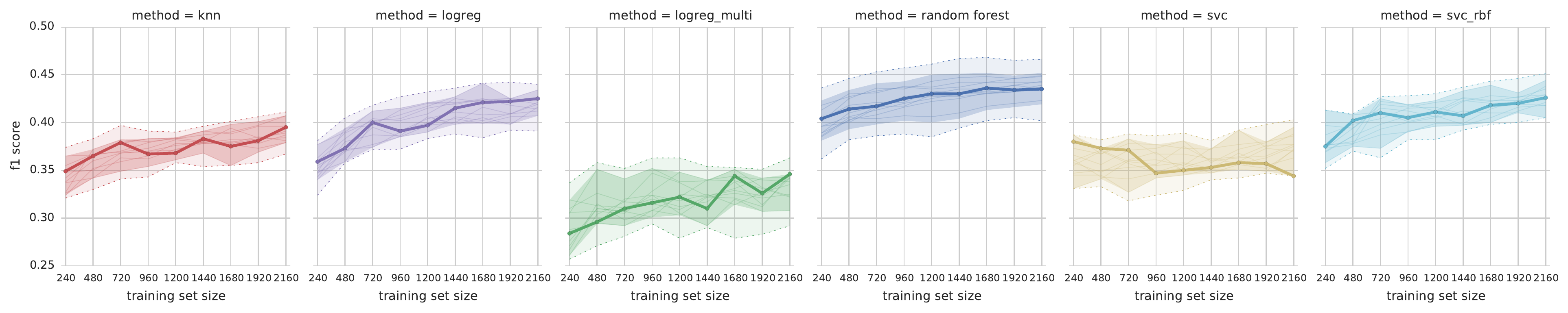}
	\caption{OAI dataset}
	\label{fig:single-oai}
\end{subfigure}
\caption{
Learning curves with $F_1$ score for models trained with different algorithms on balanced subsets of the dataset.
The dotted lines show the total max/min score for each training set size across all subsets and CV-repeats.
The solid lines (one per subset) represent elementwise median of curves for all CV-repeats.
The thick line is the elementwise median of the 11 median curves shown.
The shaded inner area contains all curves plus/minus their median average deviation (across all CV-repeats), and marks a range of the typical performance.
For exact numbers and confidence intervals see \cref{tab:performance-single-check,tab:performance-single-oai}.
}
\end{figure}

In the initial experiments on balanced subsets, the best performing algorithm was the random forest.
For the CHECK dataset, the other algorithms were competitive only at small training set sizes,
and otherwise were trailing 10\% and more behind (see \cref{fig:single-check}).
For the OAI dataset, logistic regression and SVC with the RBF kernel were closer, but on the other hand, the performance gap between random forest and the linear SVC or multi-modal regression was as large as 20\% (see \cref{fig:single-oai}).

\subsection{Performance on balanced and imbalanced training set}

\begin{figure}[!h]
\begin{subfigure}{0.48\textwidth}
	\centering
	\includegraphics[width=\textwidth]{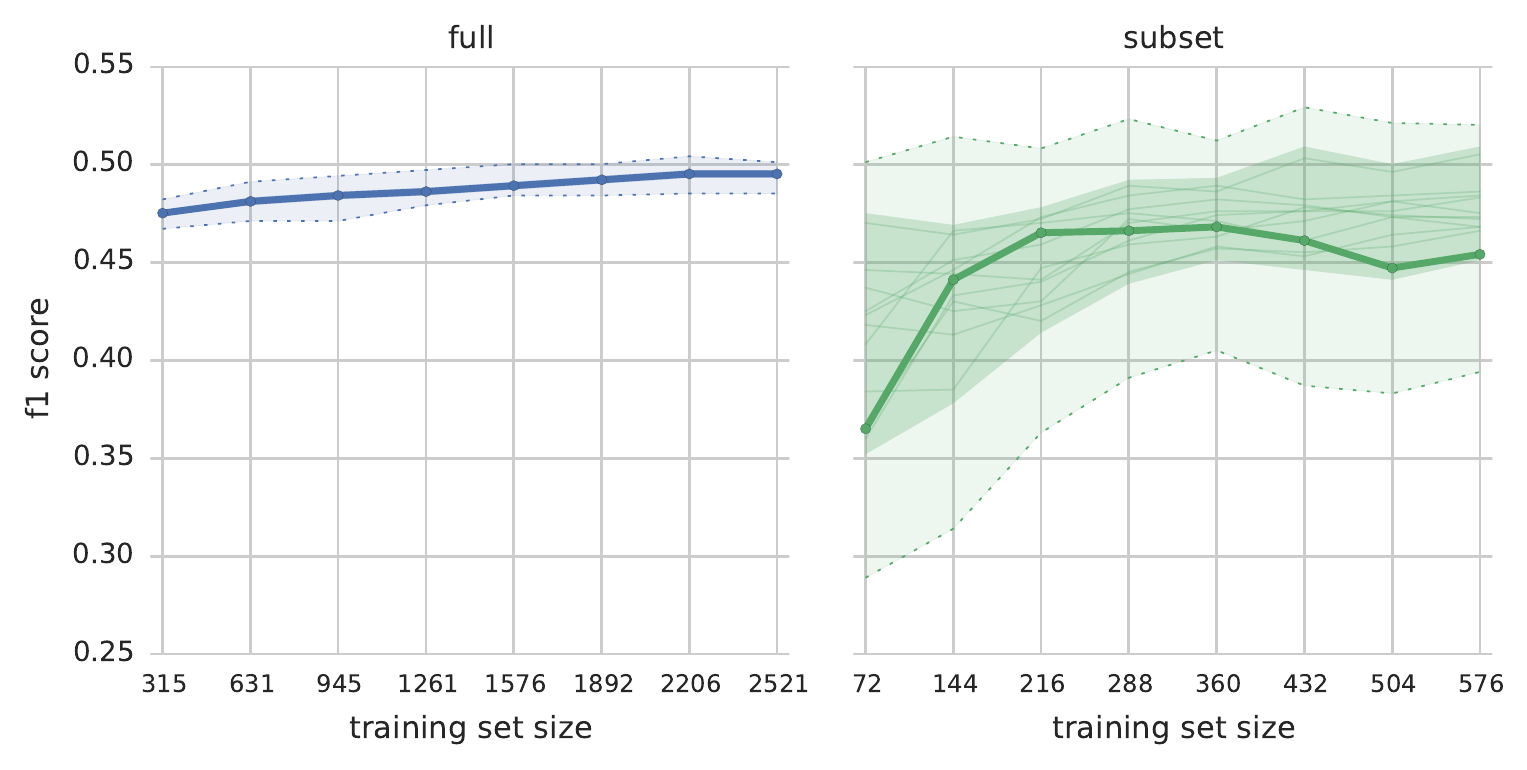}
	\caption{CHECK dataset}
\end{subfigure}
\begin{subfigure}{0.48\textwidth}
	\centering
	\includegraphics[width=\textwidth]{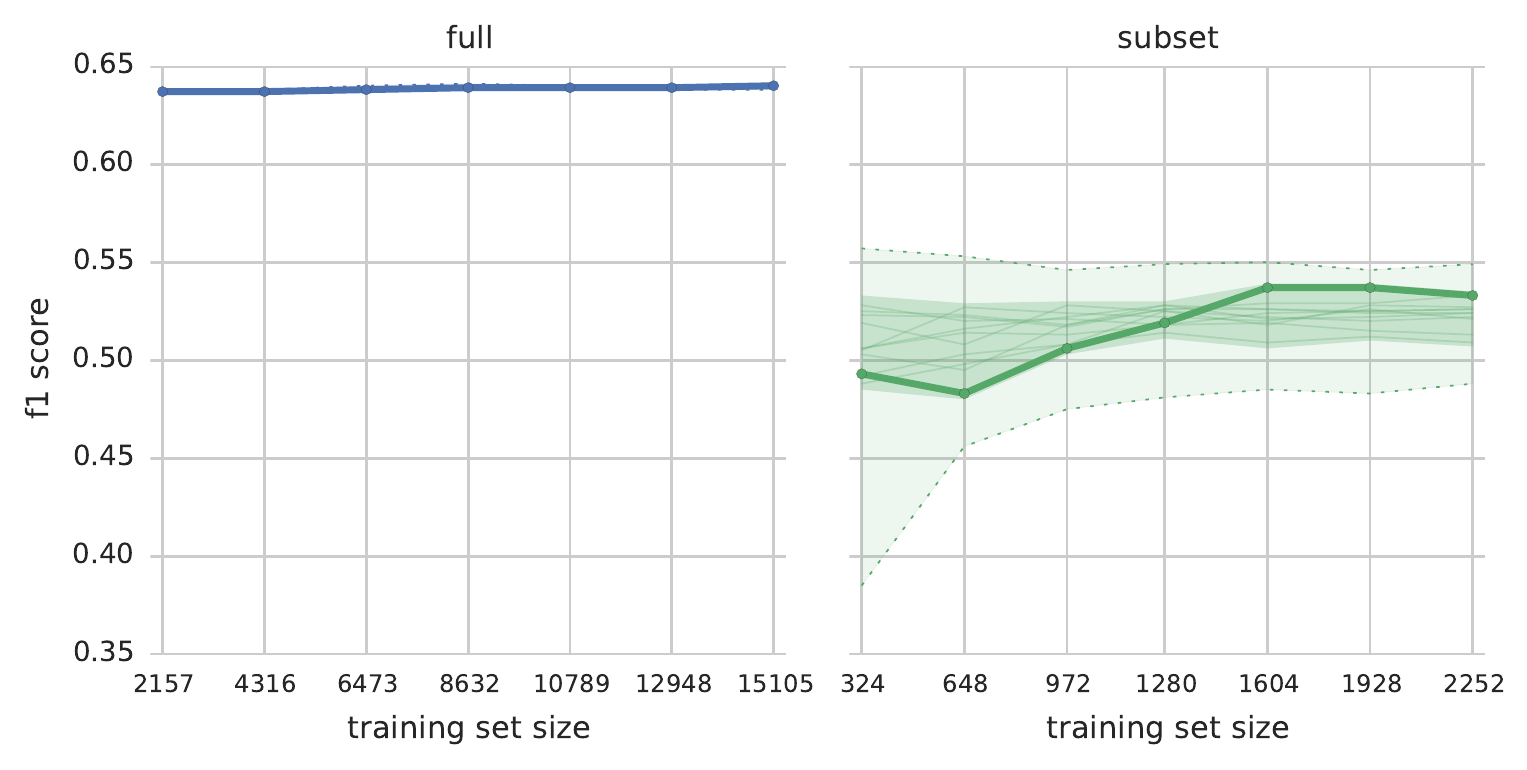}
	\caption{OAI dataset}
\end{subfigure}
\caption{
Learning curves with $F_1$ score for models trained on the full imbalanced training set (blue) or its balanced subsets (green), using the same test set.
The dotted lines show the total max/min score for each training set size.
The solid lines (one per subset) represent elementwise median of curves for all CV-repeats.
The thick line shows the median score (or elementwise median curve across subsets).
The shaded inner area represents the median average deviation (across all CV-repeats) around the median curve(s), and marks a range of the typical performance.
}
\label{fig:imbalance}
\end{figure}

\Cref{fig:imbalance} compares the performance of the cost-sensitive and balanced learning.
Two observations arise from assessing the trade-off between balanced training set and potentially easier model training, and imbalanced training set with a larger number of instances to train on.
Firstly, the bigger training set largely reduced the variance in model performance.
Secondly, the typical (median) learning curve on the full set had a higher performance at every training set size compared.
The difference was especially large in case of the OAI dataset (about 20\% in relative numbers).
Therefore, in all subsequent experiments we used the full imbalanced training set and the cost-sensitive learning.

\subsection{Performance of multi-model methods}
\label{sec:multi-model}

\begin{figure}[!htb]
\begin{subfigure}{\textwidth}
	\includegraphics[width=\textwidth]{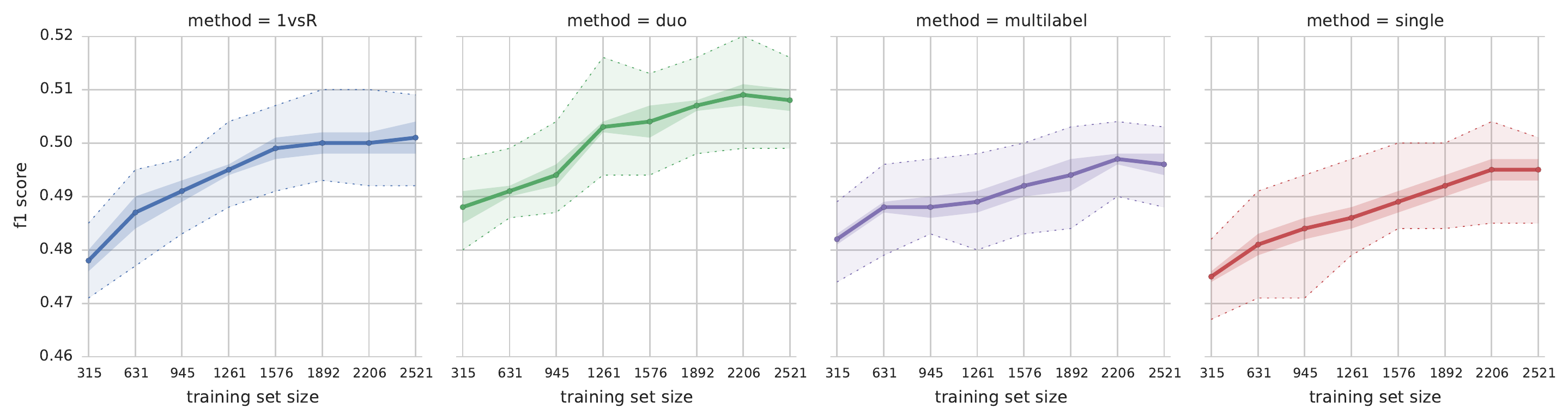}
	\caption{CHECK dataset}
	\label{fig:multi-check}
\end{subfigure}
\begin{subfigure}{\textwidth}
	\includegraphics[width=\textwidth]{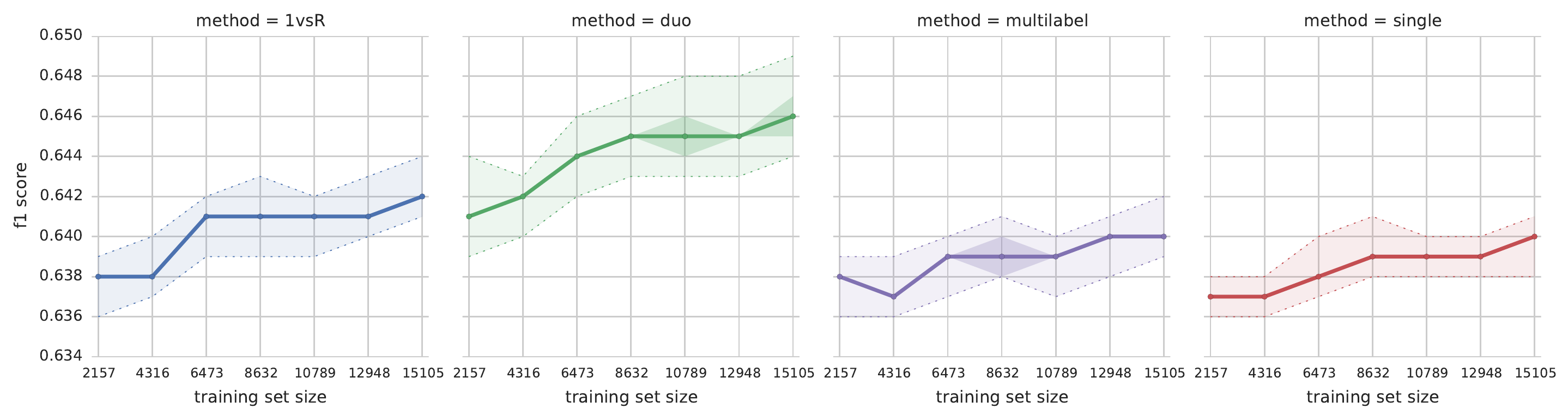}
	\caption{OAI dataset}
	\label{fig:multi-oai}
\end{subfigure}
\caption{
Learning curves with $F_1$ score for multi-model / multi-label methods trained on imbalanced dataset.
The dotted lines show the total max/min score across all CV-repeats for each training set size.
The thick solid line shows the median score.
The shaded area marks the median average deviation (across all CV-repeats) and contains $\geq 50\%$ of scores.
For exact numbers and confidence intervals see \cref{tab:performance-multi-check,tab:performance-multi-oai}.
}
\end{figure}

\Cref{fig:multi-check,fig:multi-oai} compare the performance of multi-label and multi-model strategies, to a single model 4-class random forest (indicated as ``single'').
Although all the strategies to some degree improved over the single model, the overall performance gain was minor, especially in case of the multi-label and one-vs-rest strategies.
The \textit{duo classifier} emerged as the best option, achieving a median $F_1$ score improvement of about 2\% for CHECK and 1\% for OAI.
As a result, in subsequent experiments we used the \textit{duo classifier}.

\newpage
\subsection{Random forest parameter tuning}
\label{sec:parameters}

\begin{figure}[!b]
\begin{subfigure}{0.48\textwidth}
  \centering
  \includegraphics[width=\textwidth]{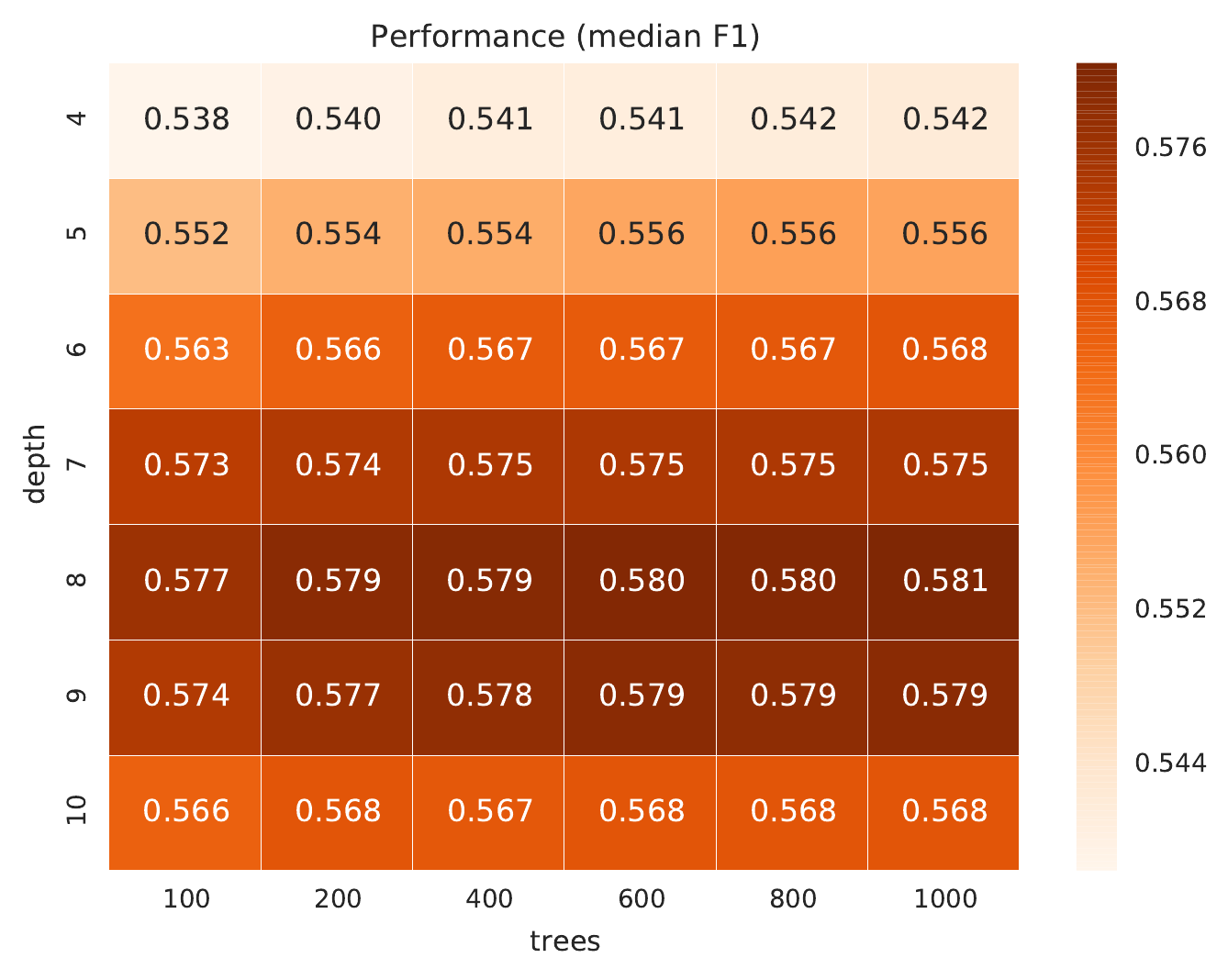}
  \caption{with \textbf{Gini impurity} criterion}
\end{subfigure}
\hfill
\begin{subfigure}{0.48\textwidth}
 \centering
 \includegraphics[width=\textwidth]{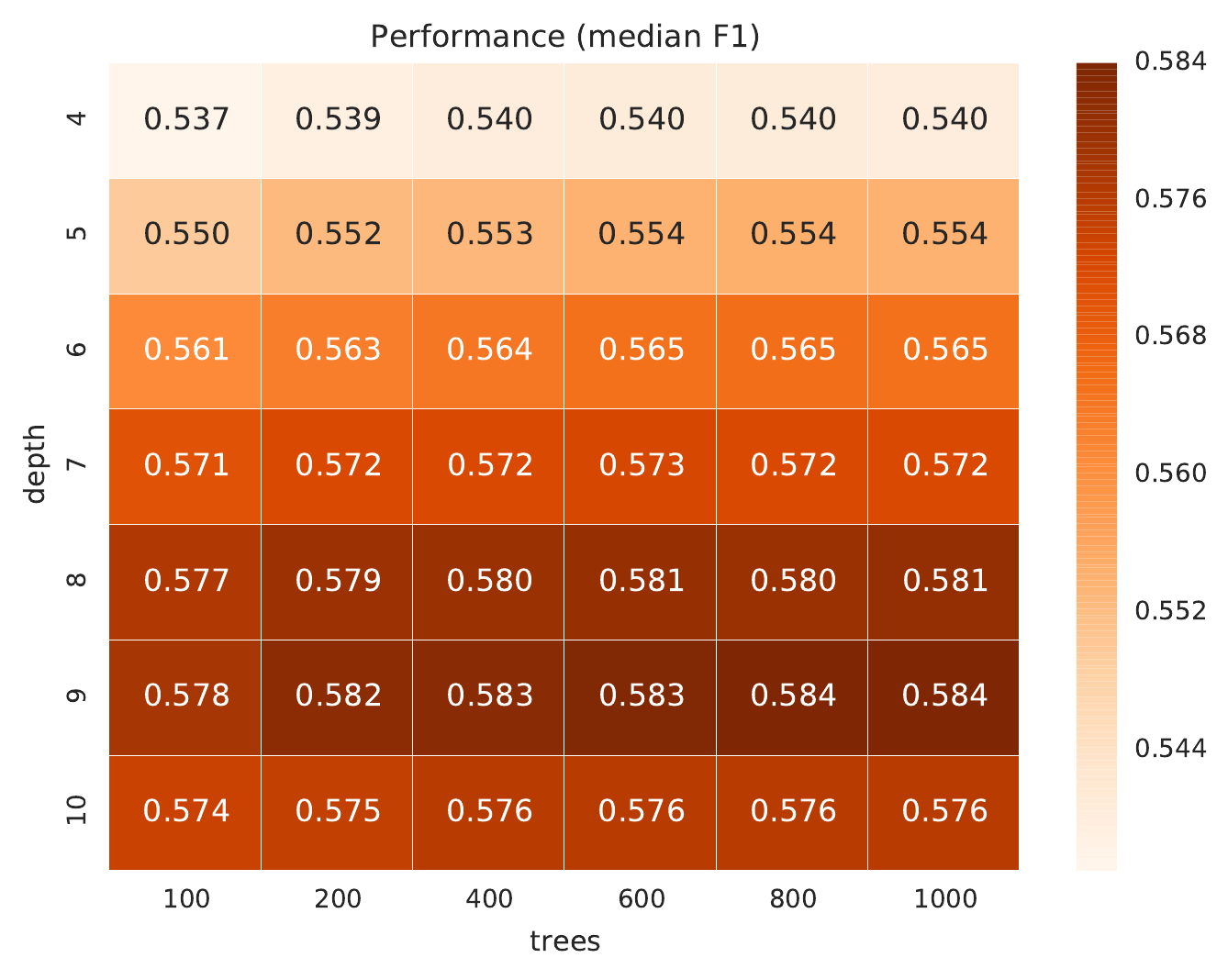}
 \caption{with \textbf{information gain} criterion}
\end{subfigure}
\caption{Performance of different configurations of the duo classifier on the \textbf{CHECK} dataset.}
\label{fig:tuning-check}
\end{figure}

\begin{figure}[!t]
\begin{subfigure}{0.48\textwidth}
  \centering
  \includegraphics[width=\textwidth]{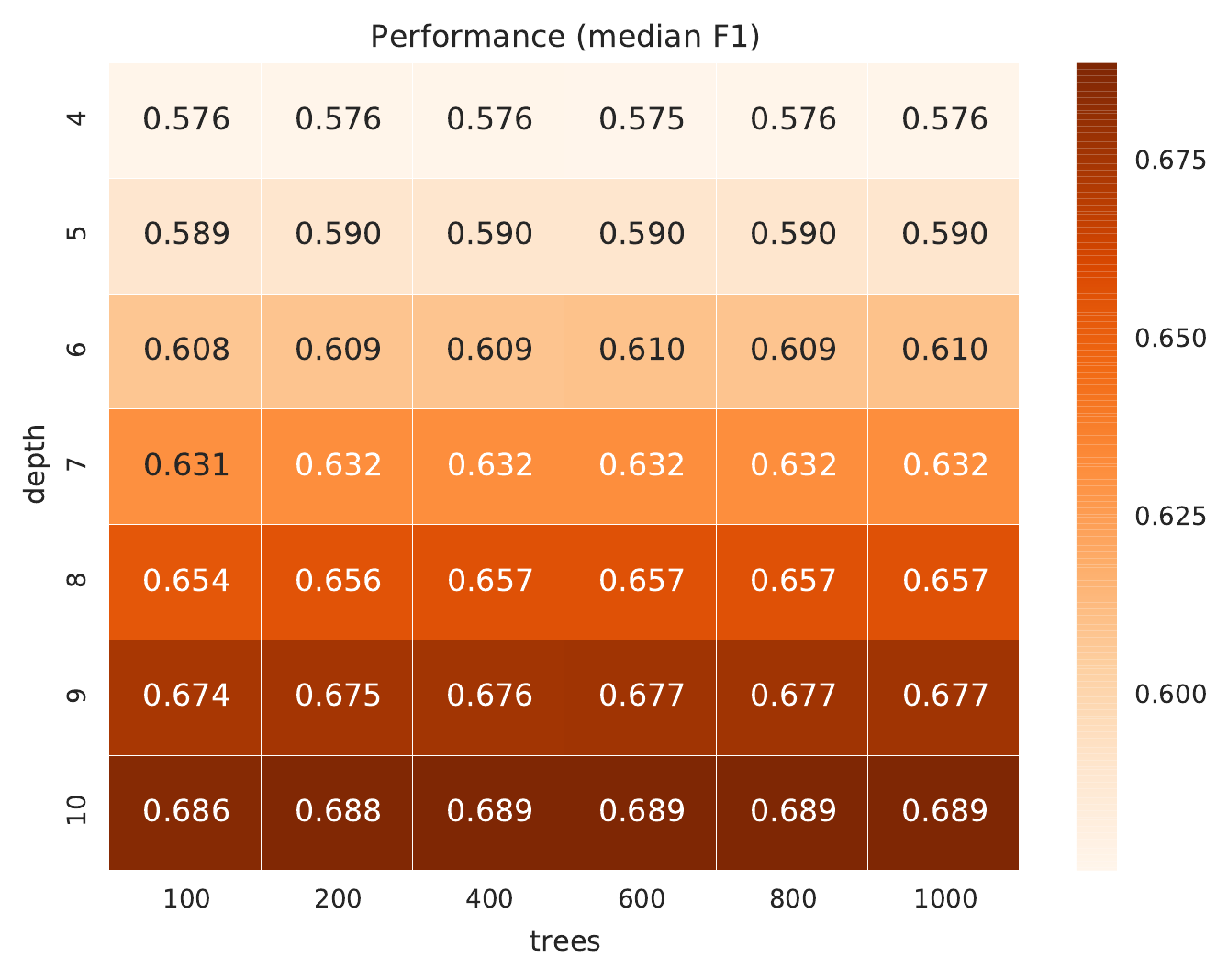}
  \caption{with \textbf{Gini impurity} criterion}
\end{subfigure}
\hfill
\begin{subfigure}{0.48\textwidth}
 \centering
 \includegraphics[width=\textwidth]{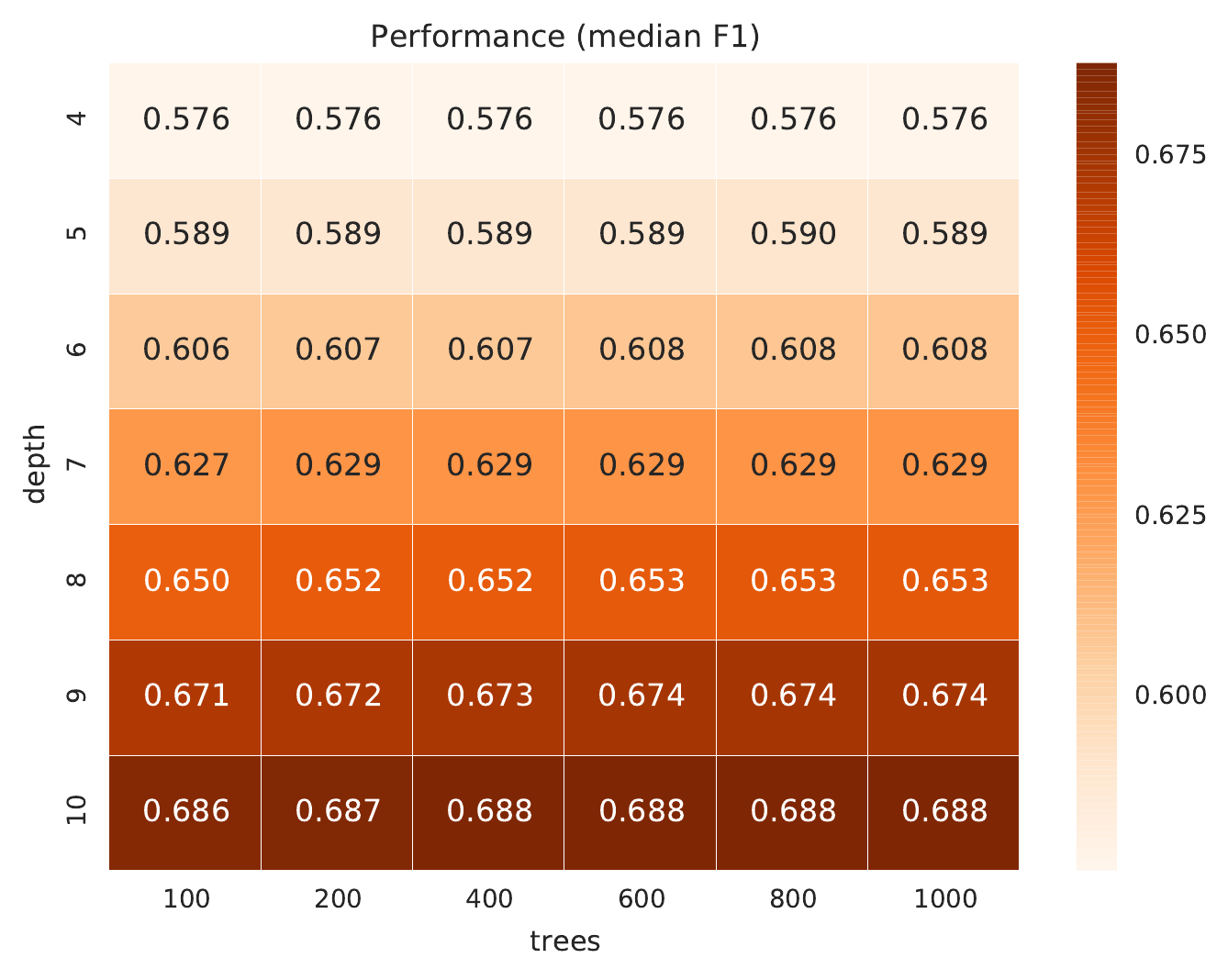}
 \caption{with \textbf{information gain} criterion}
\end{subfigure}
\caption{Performance of different configurations of the duo classifier on the \textbf{OAI} dataset.}
\label{fig:tuning-oai}
\end{figure}

\Cref{fig:tuning-check,fig:tuning-oai} show the typical performance of the \textit{duo classifier} for different algorithm configurations.
Each figure reports the $F_1$ score of the median run from the median CV-repeat.
The best performing configurations for CHECK were located in a sweet spot around 800 trees of maximum depth of 9 (for information gain criterion) and depth 8 (for Gini impurity criterion).
For OAI, we did not find a clear peak spot within the tested range of parameters.
The best performing configuration was the one with largest maximum depth of 10 and $\geq 400$ trees.
Perhaps configurations allowing for deeper trees could further improve the results.

A general conclusion is that above 400 trees the improvement in performance is very small, and a difference in the maximum tree depth has the largest impact on the score.
However, random forest is not over-training easily with more trees, and more trees can be useful (even if they do not improve performance), as they improve the reliability of the feature importance estimates.
On the other hand, with increased depth and larger trees, their interpretability decreases and there is more potential for overfitting.

In subsequent experiments we used the best performing configuration with lowest median absolute deviation, preferring lower depth and less trees in case of ties, in particular: \{800 trees, depth 9, \textit{entropy} criterion\} for CHECK, and \{1000 trees, depth 10, \textit{gini} criterion\} for OAI.

The expected performance ($F_1$ score) of the \textbf{final models} trained on all data, estimated with
the Bootstrap Bias Corrected Cross-Validation protocol (BBC-CV), was 0.584 --- 95\% CI (0.560, 0.609) for CHECK, and 0.689 --- 95\% CI (0.680, 0.698) for OAI.
For both datasets, the estimate is the same (with respect to rounding) as the score of a typical run of the best configuration (median of median runs for each CV-repeat).

\subsection{Feature selection experiments}

\Cref{tab:rfe} summarises the results of experiments with the recursive feature elimination (RFE) procedure.
As the table shows, the use of reduced set of features did not improve the model performance.
Its median score was about 2\% lower compared to configurations using all features.
We counted the frequency with which each feature was selected (out of 100 selection rounds = $10 \text{ repeats} \times 10 \text{ folds}$).
For CHECK only minimum JSW (left/right knee), WOMAC pain, WOMAC function, WOMAC total and height of the medial eminence (left/right) were selected 100\% of the time (see \cref{fig:rfe-selected} in Appendix).
For OAI this subset was much larger, 181 features were selected every time, and overlapped with CHECK features (except eminence which was not measured in OAI), therefore not much can be learned there.

\begin{table}[!htbp]
\centering
\input{tables/best_rfe.tex}
\caption{
Performance of the best model using all features vs. a subset of features found with the RFE procedure.
We report the median model score for a median CV-repeat and the 95\% confidence interval around it (from binomial distribution).
For the size of the selected subset of features, we report a range across all CV-repeats.}
\label{tab:rfe}
\end{table}

The main advantage of a smaller model (using a subset of features) is an easier interpretation, particularly with a substantial reduction to a median of just 12 features for CHECK (see \cref{fig:rfe-check} in Appendix).
It is also an advantage from the clinical perspective, as data collection is costly and sometimes less measurements could be preferred over slightly better performance.
However, it would not help much in case of the OAI models, where the median number of selected features was almost 20 times higher (see \cref{fig:rfe-oai} in Appendix).

\subsection{Features impact on model output}
\label{sec:interpretation}

\begin{figure}[!htb]
\begin{subfigure}{\textwidth}
	\includegraphics[width=0.47\textwidth]{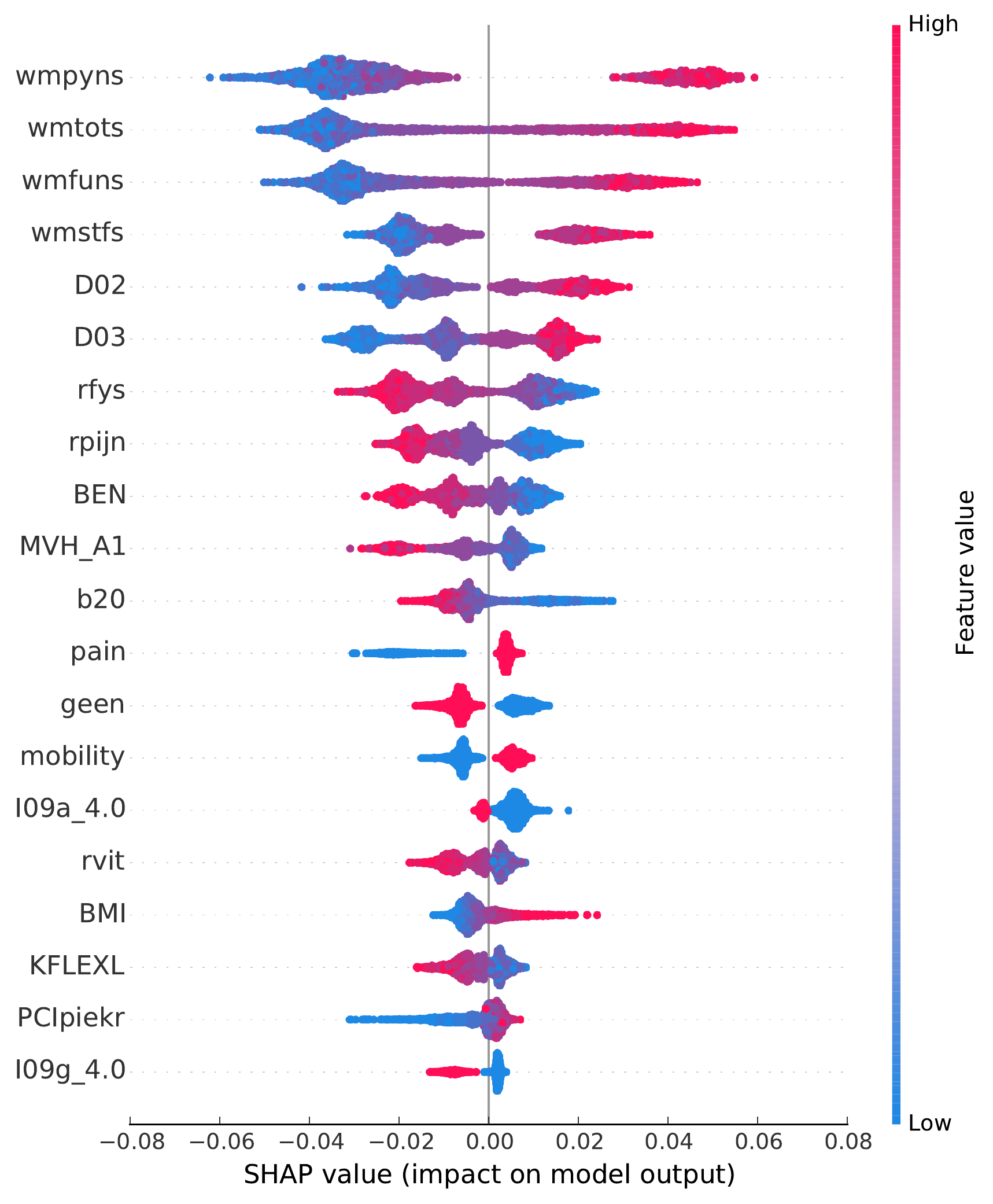}
	\hfill
	\includegraphics[width=0.51\textwidth]{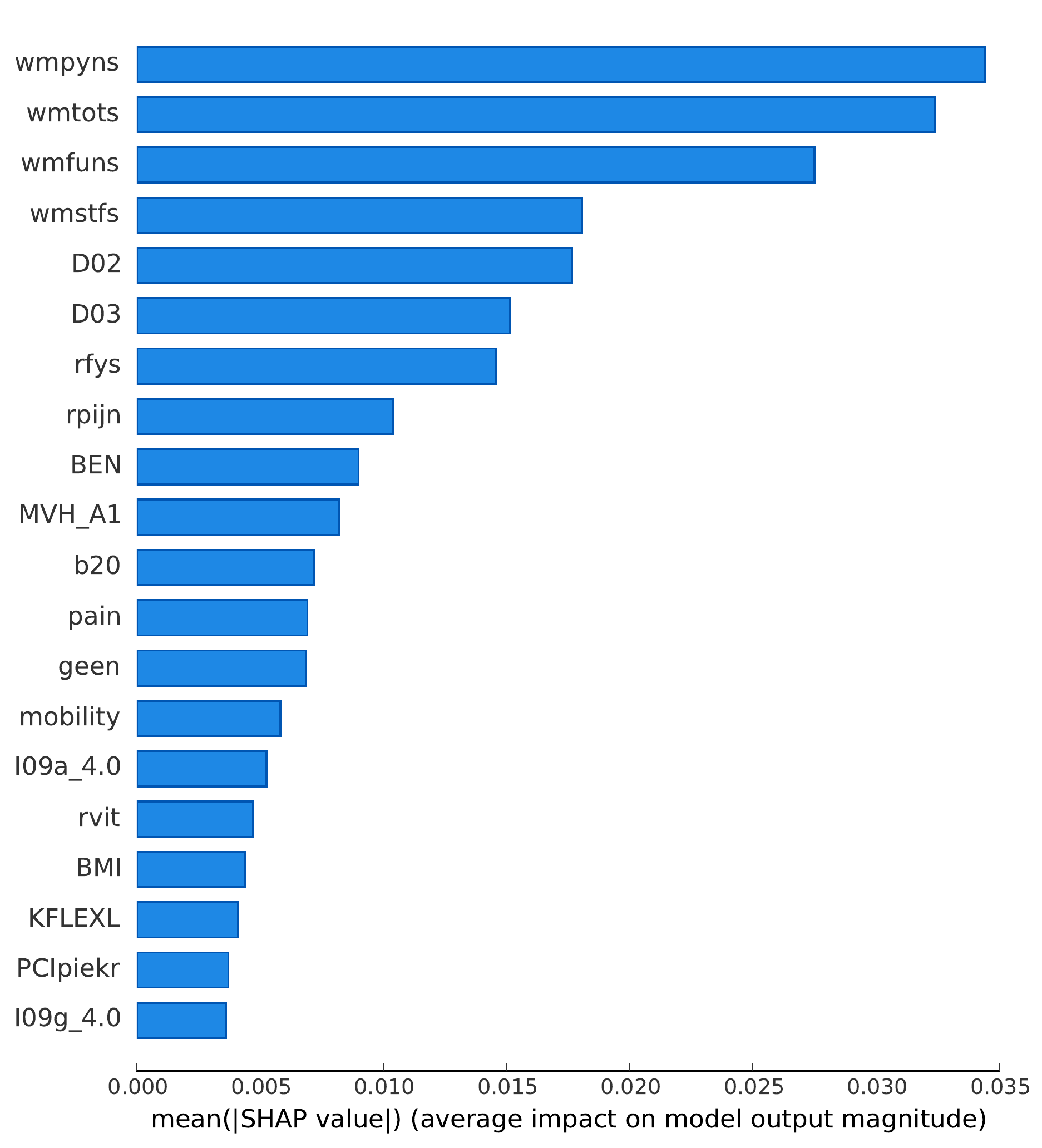}
	\caption{sub-predictor of the \textbf{P} label}
	\label{fig:shap-check-p}
\end{subfigure}
\begin{subfigure}{\textwidth}
	\includegraphics[width=0.47\textwidth]{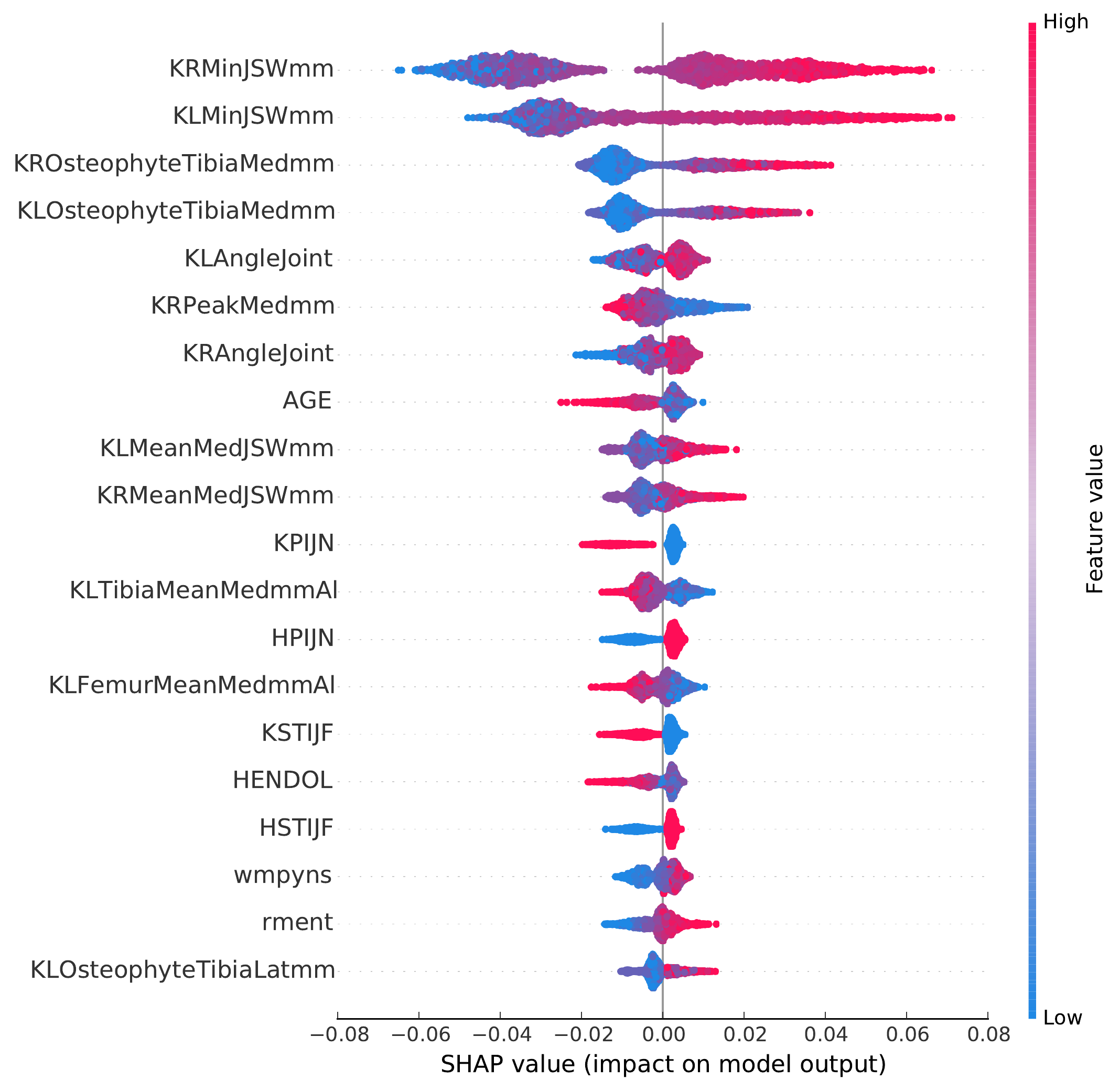}
	\hfill
	\includegraphics[width=0.51\textwidth]{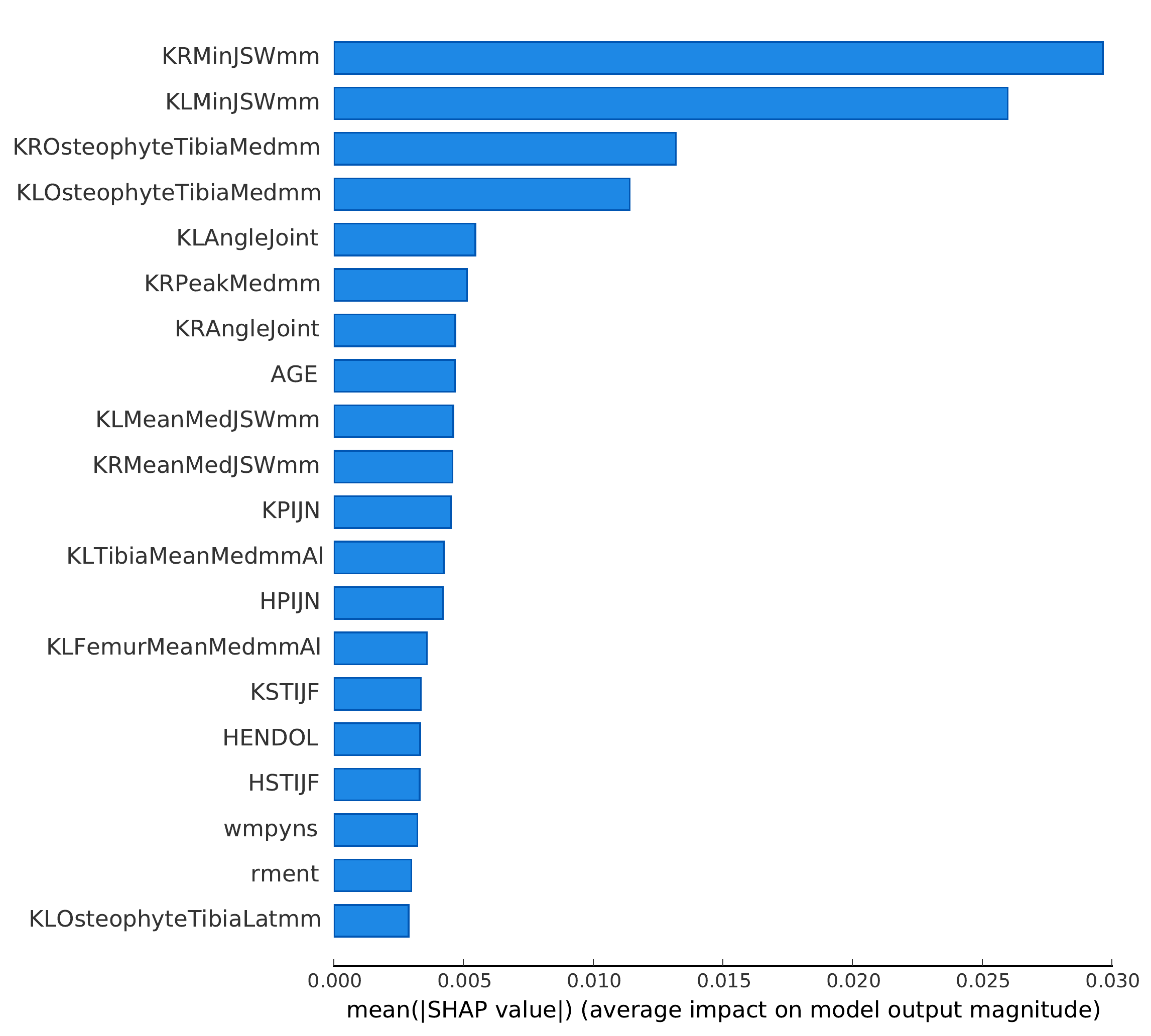}
	\caption{sub-predictor of the \textbf{S} label}
	\label{fig:shap-check-s}
\end{subfigure}
\caption{
Features impact on \textbf{P} and \textbf{S} sub-predictors output for \textbf{CHECK} dataset.
In the left panel, we show the distribution of the impact of a feature value on the model output across all instances.
A positive SHAP value indicates a positive impact (probability boost).
The colour represents the feature value (blue if low, red if high).
In the right panel, we show the average impact magnitude for all instances.
Features in both panels are ordered by their total impact.
}
\label{fig:shap-check}
\end{figure}

Although the best learning strategy was to use all features, it does not mean that they all had the same impact.
\Cref{fig:shap-check} shows features impact on the output of the \textbf{final model} trained on the entire CHECK dataset (see \cref{tab:features-check} in Appendix for feature description).
For the \textbf{P} sub-predictor, the four most impactful features are the WOMAC scores (3 sub-scores and the total score).
They all reduce the probability of assigning the \textbf{P} label if their value is low and boost that probability if their value is high (see left panel of \cref{fig:shap-check-p}).
An example of an opposite direction of influence can be seen for the \textit{rfys} feature (physical functioning from the SF-36 health survey), where higher values indicate a better health status.

For the \textbf{S} sub-predictor, the most impactful features are all related to structural degradation of the knee cartilage: the minimum JSW for both knees, with size of the osteophytes in medial tibia region and the varus angle (degree of outward bowing at the knee) further down.
Low values of minimum JSW reduce the probability of assigning the \textbf{S} label.
High values of minimum JSW, presence of large osteophytes and deviation in varus angle in range [-2.5, 0.5] boost the probability.

\begin{figure}[!htb]
\begin{subfigure}{\textwidth}
	\includegraphics[width=0.47\textwidth]{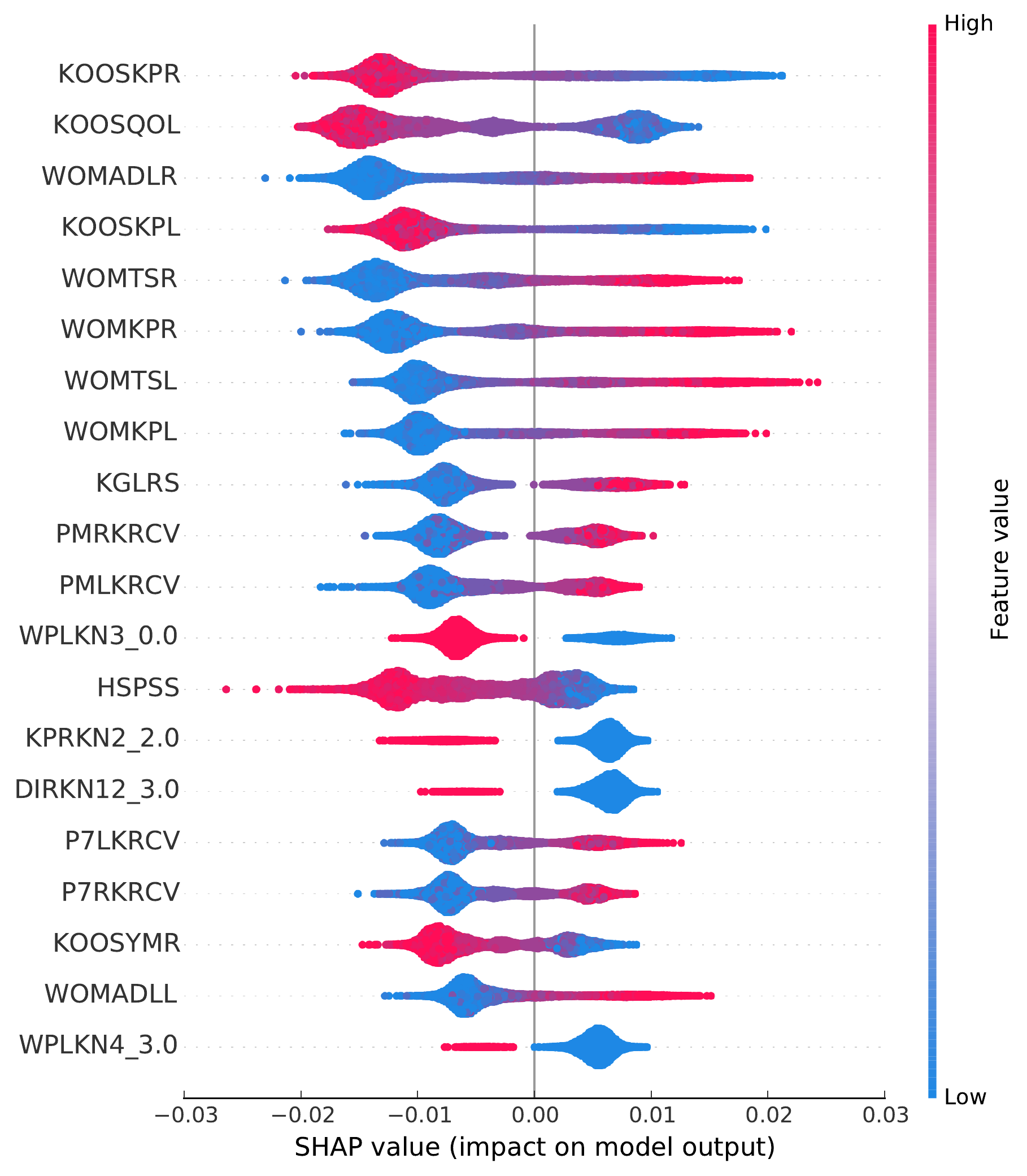}
	\hfill
	\includegraphics[width=0.51\textwidth]{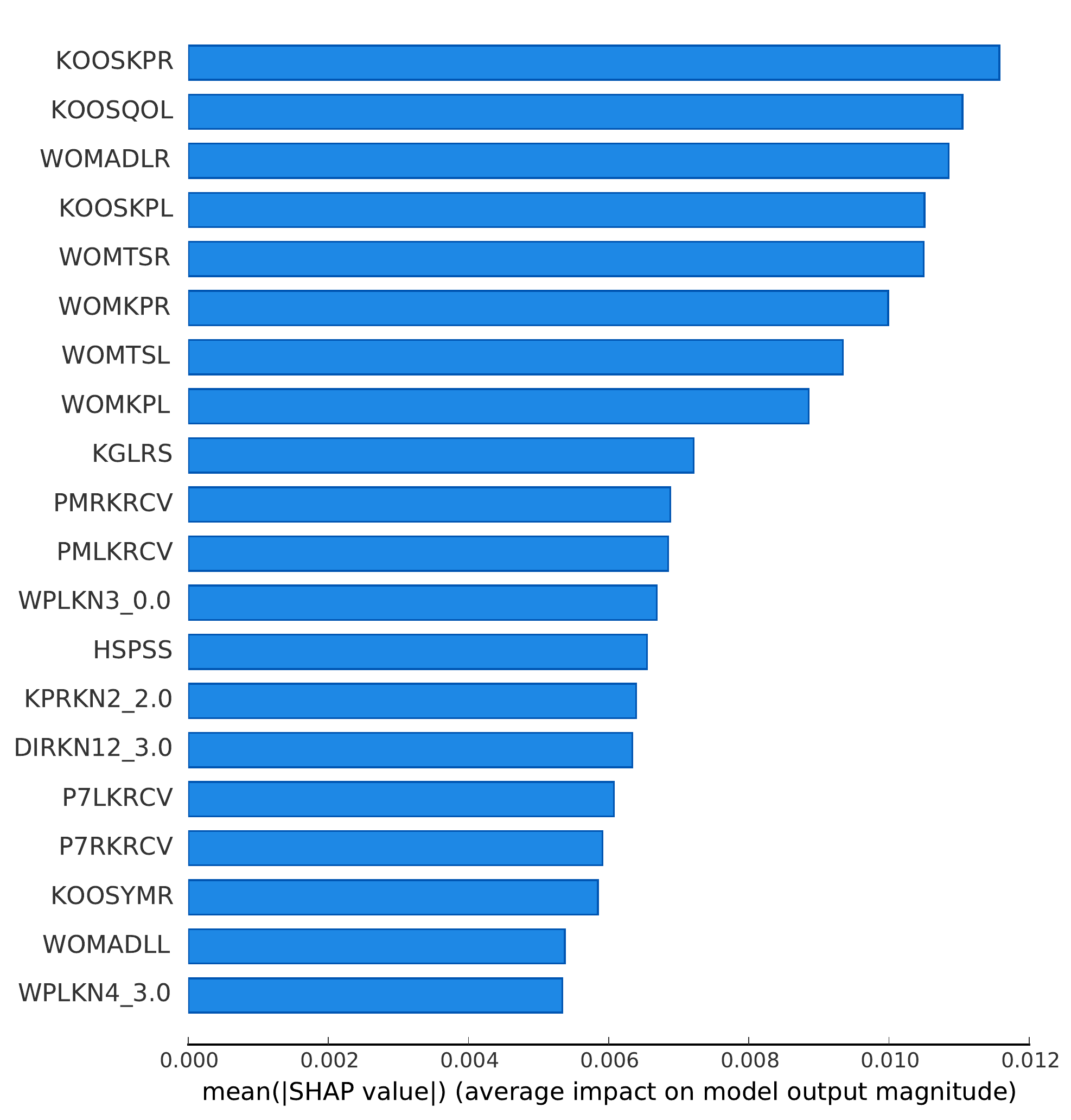}
	\caption{sub-predictor of the \textbf{P} label}
	\label{fig:shap-oai-p}
\end{subfigure}
\begin{subfigure}{\textwidth}
	\includegraphics[width=0.47\textwidth]{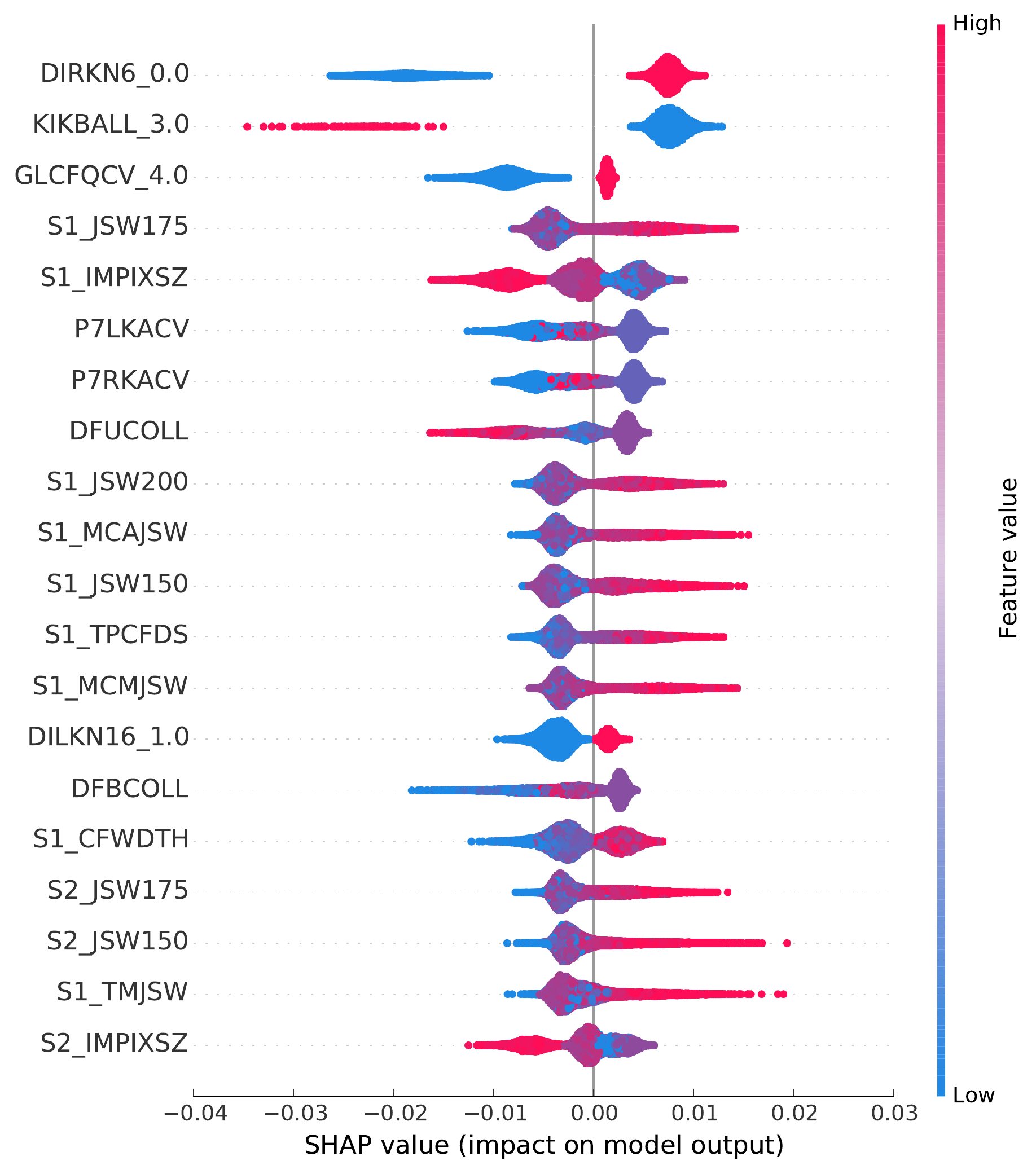}
	\hfill
	\includegraphics[width=0.51\textwidth]{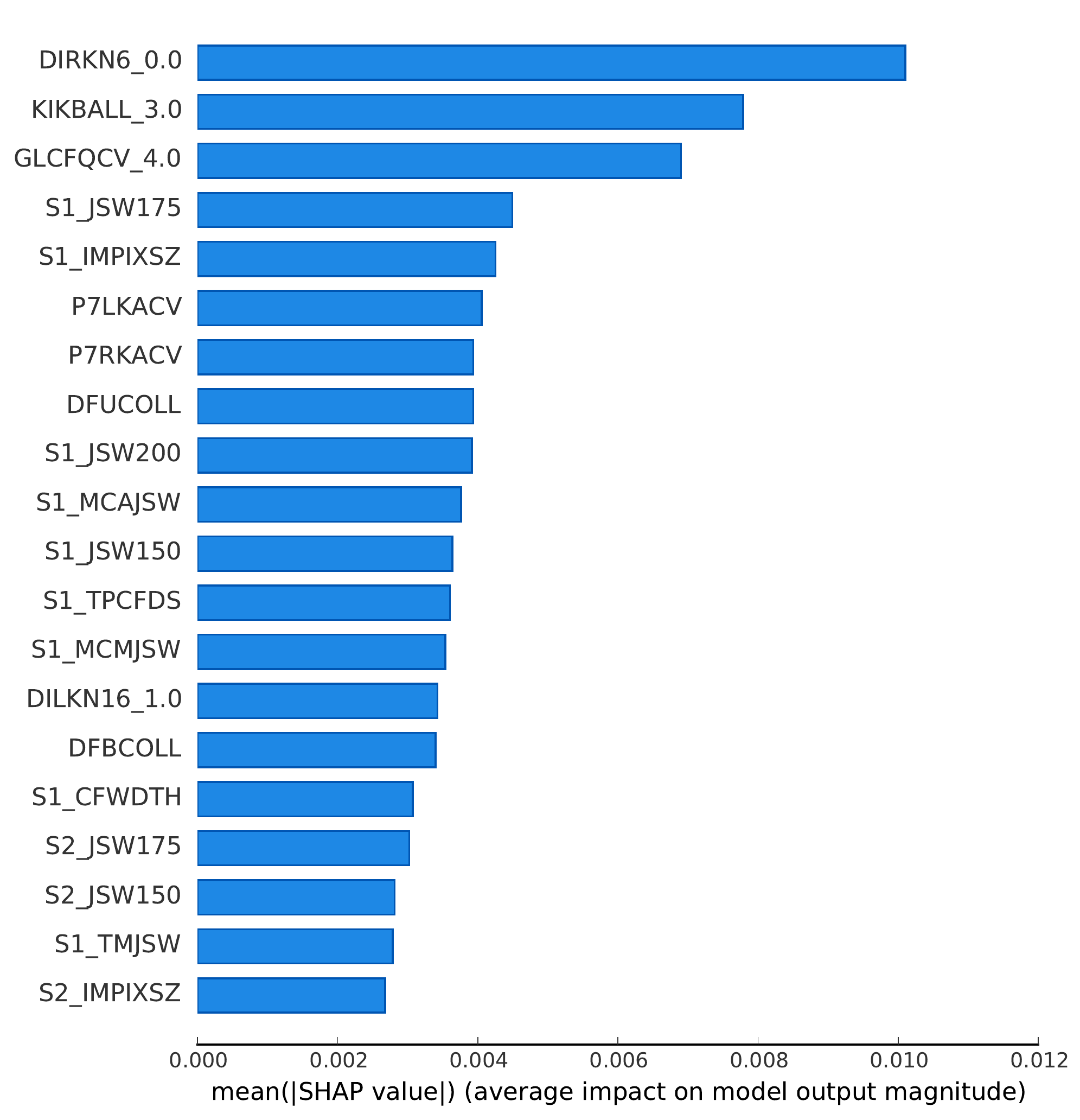}
	\caption{sub-predictor of the \textbf{S} label}
	\label{fig:shap-oai-s}
\end{subfigure}
\caption{
Features impact on \textbf{P} and \textbf{S} sub-predictors output for \textbf{OAI} dataset.
In the left panel, we show the distribution of the impact of a feature value on the model output across all instances.
A positive SHAP value indicates a positive impact (probability boost).
The colour represents the feature value (blue if low, red if high).
In the right panel, we show the average impact magnitude for all instances.
Features in both panels are ordered by their total impact.
}
\label{fig:shap-oai}
\end{figure}

\Cref{fig:shap-oai} shows the impact of features on the output of the \textbf{final model} trained on OAI dataset (see \cref{tab:features-oai} in Appendix for feature description).
For the \textbf{P} sub-predictor, the most impactful features are the KOOS and WOMAC pain scores for the left and right knee.

For the \textbf{S} sub-predictor, some of the most impactful features are pain related: \textit{DIRKN6} --- pain level while walking in the last 7 days (part of the WOMAC questionnaire), and \textit{P7RKACV} --- knee pain severity in the last 7 days.
But there are several impactful radiographic features as well, such as: \textit{JSW175} --- medial JSW at $x = 0.175mm$, \textit{MCAJSW} --- average medial JSW, or \textit{MCMJSW} --- the minimum medial JSW.
In the top 3, we can also find \textit{GLCFQCV} --- glucosamine frequency of use in past 6 months (glucosamine is a popular supplement used by OA patients).

A few features in the top make much less sense: \textit{KIKBALL} --- leg used to kick a ball, or \textit{DFUCOLL} --- difference in minutes between baseline and follow-up urine collection times, or \textit{IMPIXSZ} --- radiograph pixel size used in conversion to millimetres.
This might be a sign of attribute exploitation, as with large number of attributes in OAI and not so many instances, the model might be finding dataset specific patterns, rather than discovering general rules, and perhaps these attributes should be removed from the dataset.
Nevertheless, even if taken alone the contribution of a feature is difficult to explain, it might be useful in interaction with other features, e.g. \textit{KIKBALL\_3.0} indicates a person is ambipedal (has no dominant leg), which might trigger the use of radiographic features from both knees.

\subsection{Simulated patient selection}
\label{sec:recruitment}

We performed a selection from both datasets using the conventional clinical criteria, and compared that to two selection scenarios based on predictions of the best machine learning models: \texttt{ML-L} using the class labels, and \texttt{ML-P} using the class probabilities.
In the simpler \texttt{ML-L} scenario, we selected all instances predicted not to be in the non-progressive class (N).
In the more refined \texttt{ML-P} scenario, we selected equal number of instances most likely to be in the P+S, S or P class.

\Cref{tab:recruitment} summarise results of the selection with the conventional criteria and the \texttt{ML-L} selection scenario.
The comparison between the two revealed several issues with the conventional criteria.
Firstly, the retrieval of progressive periods was low (18\% in total) for both CHECK and OAI, especially in the \textbf{S} category (only 7\%).
Secondly, the selection  focused primarily on the \textbf{P} category, resulting in approximately half of the progressive periods from there.
On the other hand, as desired, the percentage of retrieved non-progressive periods was low (5\% for CHECK and 7\% for OAI).

\begin{table}[!htb]
\begin{subtable}{\textwidth}
	\centering
	\input{tables/expert-check.tex}
	\caption{CHECK dataset}
\end{subtable}
\vskip5mm
\begin{subtable}{\textwidth}
	\centering
	\input{tables/expert-oai.tex}
	\caption{OAI dataset}
\end{subtable}
\caption{
Subset of periods selected by the conventional clinical criteria and the \texttt{ML-L} scenario.
The number of total instances of each category is reported next to the class name.
For each category we report an absolute and relative number of included instances, and a recall percentage (how many instances of that category have been retrieved).
The ``not N'' column shows the summarised recall percentage for all progressive instances.
}
\label{tab:recruitment}
\end{table}

The \texttt{ML-L} selection scenario retrieved over 2 times more progressive periods ($\approx 45\%$ in total).
In the \textbf{S} category the retrieval was 5 times higher than the conventional criteria result.
The balance between the categories has improved for CHECK where \textbf{P} and \textbf{S} categories only differed by 2 p.p., but not for OAI, where the \textbf{S} category became dominant.
Overall, we see that our machine learning models were less conservative (i.e. have made more non-N predictions) than the conventional criteria, which resulted in retrieving more progressive instances, at the cost of incorporating higher relative percentage of non-progressive ones.

Although in the \texttt{ML-L} scenario, the machine learning had some advantages in recall levels over the conventional criteria, it selected a larger number of non-progressive instances.
It also selected 2.5--3 times more instances overall.
To make a more direct comparison, in the \texttt{ML-P} scenario we selected the same total number of instances as obtained with the conventional criteria.
The selection prioritised the instances more likely to progress and directly used the probabilities provided by the classifier.

\Cref{tab:recruitment-probabilities} shows the results of the \texttt{ML-P} selection scenario.
Not only did it reduce the number of non-progressive instances compared to the conventional criteria (by $\approx 20\%$ for CHECK and $\approx 25\%$ for OAI), but it also increased the balance between the progressive categories (boosting selection from S and P+S, while reducing the bias towards P).

\begin{table}[!htb]
\centering
\input{tables/expert-probabilities.tex}
\caption{Comparison between selection with conventional clinical criteria and the \texttt{ML-P} scenario.}
\label{tab:recruitment-probabilities}
\end{table}

\section{Discussion}
\label{sec:discussion}

We hypothesised that machine learning models predicting OA progression could be used to select fast progressing patients more effectively than the conventional inclusion criteria.
In a search for the most performant learning process configuration, we used a careful evaluation focused on the median performance.
For statistical stability of the results, we used repeated cross-validation and trained multiple models for each fold using different random seeds.
We found random forest to stand out as the best learning algorithm.
The cost-sensitive learning with random forest outperformed the balanced learning on down-sampled training set, and reduced the variance in model scores.
The multi-model approach with the \emph{duo classifier} further improved the results.
Contrary to our expectations, we did not obtain better models with recursive feature elimination.

When predictions of the best models were used to simulate patient selection, we observed a substantial reduction in the number of undesired non-progressive cases.
This findings could impact the future clinical trials design, and potentially improve their efficiency.
A machine learning model similar to ours, could be applied to the screening data during the inclusion phase of a trial, and suggest which patients should be enrolled in the study.
The screening visits could be continued until the trial is sufficiently enriched with patients who are likely to show disease progression within the trial period, and allow for more effective treatment evaluation.

\subsection{Limitations and future work}

A clear limitation of the experiment design, was the weak preprocessing strategy for the OAI dataset.
We did not identify the ordinal attributes and therefore we applied one-hot encoding to every categorical attribute regardless of its semantics.
A similar problem repeated for the continuous attributes with low number of unique values, which were treated as categorical and unnecessarily encoded.
This led to a construction of less general decision trees, with splits relying on specific attribute values (rather than value ranges), and made the model less trustworthy from a clinical point of view.

A related issue is the clinical relevance of the features the models relied on.
It is inevitable that some of the features will be exploited to make shortcut decisions, despite not representing any real knowledge.
For that reason, it is important to look ``inside'' the models and iteratively refine the data representation in the training set, to gradually eliminate the potential for misuse.
But this process is not trivial, as models can use hard to explain features (indirectly associated with progression) as a proxy for what is not directly observed.
Although we eliminated some of the feature misuse already (e.g. our first OAI models were misusing the image barcodes), still more work needs to be done in this regard, involving further dialogue with the domain experts.

In terms of further improvement of the model performance, it might be possible to achieve better results if the configuration of parameters used to train the \textit{duo classifier} is not shared between its sub-classifiers.
That is, each of the sub-classifiers could have been tuned separately, including a dedicated feature elimination procedure (perhaps even with more inner cross-validation folds), to maximise its individual performance.
Whether that would lead to a better overall performance is a matter of experiment, as it might as well increase the risk of over-training.
For certain, it would require a substantial additional computational effort --- the longest RFE experiment we performed so far, already took over 200 CPU days on our HPC cluster (using Intel Xeon E5-2690 processor).
Moreover, due to the sequential nature of the RFE procedure (features were eliminated one by one), it cannot be easily sped up through parallelisation.

Another question is, how easy would it be to implement our approach in clinical practice.
The main obstacle would be the process of patients' data collection.
It is usually performed on a rolling basis (over the course of several months), due to logistics reasons (e.g. limited access to equipment or personnel), which makes a single selection step, as we performed in this work, impractical.
Therefore, further work is needed on extending this approach towards a multi-step selection, in which decisions are made on small batches of patients as their data become available, without sacrificing the overall selection quality.

\subsection{Choice of performance measure}
\label{sec:discussion-measures}

In this work, inspired by the similarity of the patient selection problem to the task of document retrieval, we decided to measure the classification performance with $F_1$ score.
Below, we briefly discuss the advantages and drawbacks of several alternative measures.

Area under the ROC curve (AUC) is commonly used in medical binary classification tasks such as cases vs. controls analysis.
Although a generalisation to multi-class problems, \textit{M-score}, has been proposed by Hand and Till \cite{Hand2001}, the use of AUC for model comparison has been strongly criticised by Hand himself.
He not only pointed out problems with comparison of the crossing ROC curves (where difference in AUC creates false impression that one curve dominates the other), but also demonstrated the measure incoherence \cite{Hand2009} (AUC evaluates different classifiers with a different metric, as it depends on the score distributions, which depend on the classifier).
Hand proposed \textit{H-measure} as a replacement for the AUC, but it has been only defined for binary classification.

Matthew's Correlation Coefficient (MCC) is another measure of binary classification performance \cite{Matthews1975} that has been extended to handle multi-class problems \cite{Gorodkin2004}.
Its main merit is in taking into account true negatives (accuracy or $F_1$ do not), which makes MCC especially useful when negative examples are the minority.
Unfortunately, this is not the case in the patient selection task.

Measures based on the error matrix (like $F_1$ score or MCC), do not take into account the distance in the class probability space (they treat every mistake the same, regardless of its scale).
There are several measures that do, but they lack in other aspects.
For example, area under the precision-recall curve (AUPRC) does not generalise to a multi-class case.
Log-loss or the Brier distance can handle multi-class problems, but they do not address the class imbalance directly.
Perhaps the patient selection task would benefit from a dedicated measure of performance designed to align with the specific recruitment requirements.

\subsection{Related work}

Although several long-term OA clinical studies have been completed and their outcomes analysed in detail, very little research has been done on improving the patient selection process.
To our best knowledge, this work is a first attempt at building machine learning models that can compete with the established clinical practice.

Our approach differs from most of the analyses found in the literature in two important ways.
Firstly, it does not focus on determining the risk factors, but on the prediction of the disease progression.
Secondly, it defines the progression within a strict time window and targets the change in fine-grained radiographic measurements (JSW), rather than just a categorical difference in the KL/JSN grade.

Most of the previous works do not focus on disease progression, but analyse OA incidence instead, where a patient can either be diagnosed with OA (typically when KL grade $\geq 2$) or be ``OA free'' (when KL grade $\leq 1$).
The incidence of disease is then defined, as a change in diagnosis of the same knee between the baseline and the follow-up visit, and is analysed with statistical methods to determine the risk factors (usually odds ratios with univariate analysis of variance, or multivariate logistic regression).
Some authors go a bit further and test the logistic regression models on a binary classification task (cases vs. controls) \cite{Zhang2011,Kinds2012,Kerkhof2014} hand-picking the input variables.
However, as Jamshidi~et~al. point out in their recent perspective article \cite{Jamshidi2018}, very few authors reach beyond statistical analysis and build machine learning models.

Yoo~et~al.\cite{Yoo2016} trained an artificial neural network with 7 inputs and 3 hidden layers to directly predict the KL grade, obtaining AUC $> 0.8$.
However, they only focused on discriminating between KL grade levels at baseline, rather than trying to predict future disease progression.
Similar results were obtained with random forest by Minciullo~et~al.\cite{Minciullo2017} who were able to discriminate between cases and controls with AUC $> 0.85$, but in the prediction task (same cohort, OA incidence after 84 months) achieved a much lower score ($\approx 0.6$).
Better OA incidence prediction (AUC $> 0.8$) was reported by Lazzarini~et~al.\cite{Lazzarini2017} who used random forest with an iterative feature elimination heuristic (RGIFE).

These results are not directly comparable, as the models were trained on data from different cohorts.
As a consequence, the models operated on a different input, and used inconsistent definition of the outcome (the OA incidence was defined over a period of varying length: 10 \cite{Kerkhof2014}, 7 \cite{Minciullo2017}, or 2.5 \cite{Lazzarini2017} years).
Moreover, due to the AUC measure incoherence discussed earlier, any comparison between these models would be, at most, approximate.

When it comes to the definition of the progression used in this article, in many aspects it is similar to the definition used by the FNIH OA Biomarkers Consortium (e.g. \cite{Eckstein2015, Kraus2017}).
They likewise defined four categories of patients (N, P, S, and P+S) based on the change in WOMAC/JSW over time, but flexibly allowed the progression to happen at 2, 3 or 4 year follow-up.
In contrast to our fixed 2 year time period, this does not select for a fast progression.
Furthermore, the analysis performed in these works, is again focused on the risk factors only.
In the best case, a test of discriminatory power is performed (without correcting for overfitting) but no independent prediction is attempted.
Notable exception is the work by Hafezi-Nejad~et~al.\cite{Hafezi-Nejad2017} who used a small artificial neural network with 10 inputs and 1 hidden layer to predict the joint space loss, and with a single training/test set random split and 100 runs, obtained an average AUC of 0.669.

\section{Conclusions}


The aim of this work has been to test if the machine learning models can be more predictive of the future knee OA progression than the conventional clinical selection criteria.
We focused on a short progression time window typical for clinical trials.
Using data from two long-term knee OA studies (CHECK and OAI), we experimented with different learning strategies to build the final models, and obtained the best results with a custom-made \textit{duo classifer}.
The model-based selection, compared to the conventional criteria, resulted in 20--25\% less non-progressive instances and more balanced retrieval of the progressive ones.

These results put into question the effectiveness of the conventional selection criteria, which although straightforward to apply in practice, were found to be less predictive of the future disease progression.
At the same time, these results reveal a potential to develop more precise screening tools, leading to better designed clinical trials, and in consequence, to more successful evaluation of therapies, which is important for patients, scientific community, pharmaceutical industry and the ageing society in general.

Further work is needed before this potential is fully understood.
Our approach needs to be implemented into the clinical practice, and tested in a real study.
That involves a number of challenges, from methodology of the model evaluation to logistics of the selection process.
We hope to solve some of them in the APPROACH study recruitment process, and based on its future results, assess the practical impact of the model-based selection.

\section*{Contributions}

This section describes the roles of all contributors (whether formally listed as authors or named in acknowledgements) using the CRediT taxonomy \cite{Brand2015}.
\vskip1mm

{\small
\textbf{Jaume Bacardit}: Conceptualisation, Methodology, Resources, Writing -- Original Draft, Supervision, Project Administration, Funding Acquisition.
\textbf{Anne-Christine Bay-Jensen}: Writing - Review \& Editing, Funding Acquisition.
\textbf{Francis Berenbaum}: Writing - Review \& Editing.
\textbf{Janneke Boere}: Project Administration.
\textbf{Ida Haugen}: Writing - Review \& Editing.
\textbf{Leonie Hussaarts}: Project Administration.
\textbf{Margreet Kloppenburg}: Writing - Review \& Editing.
\textbf{Christoph Ladel}: Conceptualisation, Supervision, Project Administration, Funding Acquisition.
\textbf{Floris Lafeber}: Conceptualization, Writing - Review \& Editing, Funding Acquisition.
\textbf{Jonathan Larkin}: Conceptualization, Writing - Review \& Editing, Project Administration, Funding Acquisition.
\textbf{Marieke Loef}: Writing - Review \& Editing.
\textbf{John Loughlin}: Conceptualisation, Resources, Supervision, Project Administration, Funding acquisition.
\textbf{Anne Marijnissen}: Writing - Review \& Editing.
\textbf{Ali Mobasheri}: Conceptualization, Funding Acquisition.
\textbf{Sjaak Peelen}: Writing - Review \& Editing.
\textbf{Florence Petit Dop}: Conceptualization, Writing - Review \& Editing, Funding Acquisition.
\textbf{Jérémie Sellam}: Writing - Review \& Editing.
\textbf{Erwin van Spil}: Writing - Review \& Editing.
\textbf{Harrie Weinans}: Conceptualization, Funding Acquisition, Project Administration.
\textbf{Paco Welsing}: Conceptualisation, Methodology, Writing -- Review \& Editing.
\textbf{Paweł Widera}: Conceptualisation, Methodology, Software, Formal Analysis, Investigation, Writing --- Original Draft, Visualisation.
}

\section*{Acknowledgements}

We thank Janet Wesseling for providing explanations of the CHECK codebook, Anne-Christine Bay-Jensen, Francis Berenbaum, Ida Haugen, Marieke Loef, Anne Marijnissen, Margreet Kloppenburg, Sjaak Peelen, Jérémie Sellam and Erwin van Spil for comments and suggestions on the draft of this manuscript, and Janneke Boere and Leonie Hussaarts for coordination of the research activity.

The research leading to these results has received support from the Innovative Medicines Initiative Joint Undertaking under Grant Agreement no.\textbf{115770}, resources of which are composed of financial contribution from the European Union's Seventh Framework Programme (FP7/2007-2013) and
EFPIA companies’ in kind contribution. See \url{http://www.imi.europa.eu/} and \url{www.approachproject.eu/}.

This research used the High Performance Computing cluster at the School of Computing at Newcastle University.

\section*{Disclaimer}

This communication reflects the views of the authors and neither IMI nor the European Union and EFPIA are liable for any use that may be made of the information contained herein.

\section*{Data availability}

The data from machine learning experiments performed during this study are available under a CC0 licence at \href{https://doi.org/10.25405/data.ncl.10043060}{doi:10.25405/data.ncl.10043060}.

The CHECK and OAI cohorts are controlled access datasets available from their owners at \href{https://doi.org/10.17026/dans-252-qw2n}{doi:10.17026/dans-252-qw2n} and \url{https://oai.epi-ucsf.org/}.

\footnotesize
\bibliographystyle{ieeetr-url}
\bibliography{references}

\newpage
\appendix
\section{Appendix}

\renewcommand\thefigure{A\arabic{figure}}
\renewcommand\thetable{A\arabic{table}}
\setcounter{figure}{0}
\setcounter{table}{0}

\begin{table}[h]
\centering \small
\input{tables/training-check.tex}
\caption{
Comparison of algorithm performance on balanced subsets of the \textbf{CHECK} dataset (corresponding to \cref{fig:single-check}).
We report the median F1-score and confidence intervals around median (from binomial distribution) across all subsets and CV-repeats, for selected training set sizes (3/9, 6/9, and 9/9).
}
\label{tab:performance-single-check}
\end{table}

\begin{table}[h]
\centering \small
\input{tables/training-oai.tex}
\caption{
Comparison of algorithm performance on balanced subsets of the \textbf{OAI} dataset (corresponding to \cref{fig:single-oai}).
We report the median F1-score and confidence intervals around median (from binomial distribution) across all subsets and CV-repeats, for selected training set sizes (3/9, 6/9, and 9/9).
}
\label{tab:performance-single-oai}
\end{table}

\begin{table}[h]
\centering \small
\input{tables/training-multi-check.tex}
\caption{
Comparison of performance of multi-model / multi-label methods trained on the \textbf{CHECK} dataset (corresponding to \cref{fig:multi-check}).
We report the median F1-score and confidence intervals around median (from binomial distribution) across all CV-repeats, for selected training set sizes (3/8, 5/8, and 8/8).
}
\label{tab:performance-multi-check}
\end{table}

\begin{table}[h!]
\centering \small
\input{tables/training-multi-oai.tex}
\caption{
Comparison of performance of multi-model / multi-label methods trained on the \textbf{OAI} dataset (corresponding to \cref{fig:multi-oai}).
We report the median F1-score and confidence intervals around median (from binomial distribution) across all CV-repeats, for selected training set sizes (3/7, 5/7, and 7/7).
}
\label{tab:performance-multi-oai}
\end{table}

\begin{figure}[h]
	\centering
	\includegraphics[width=0.7\textwidth]{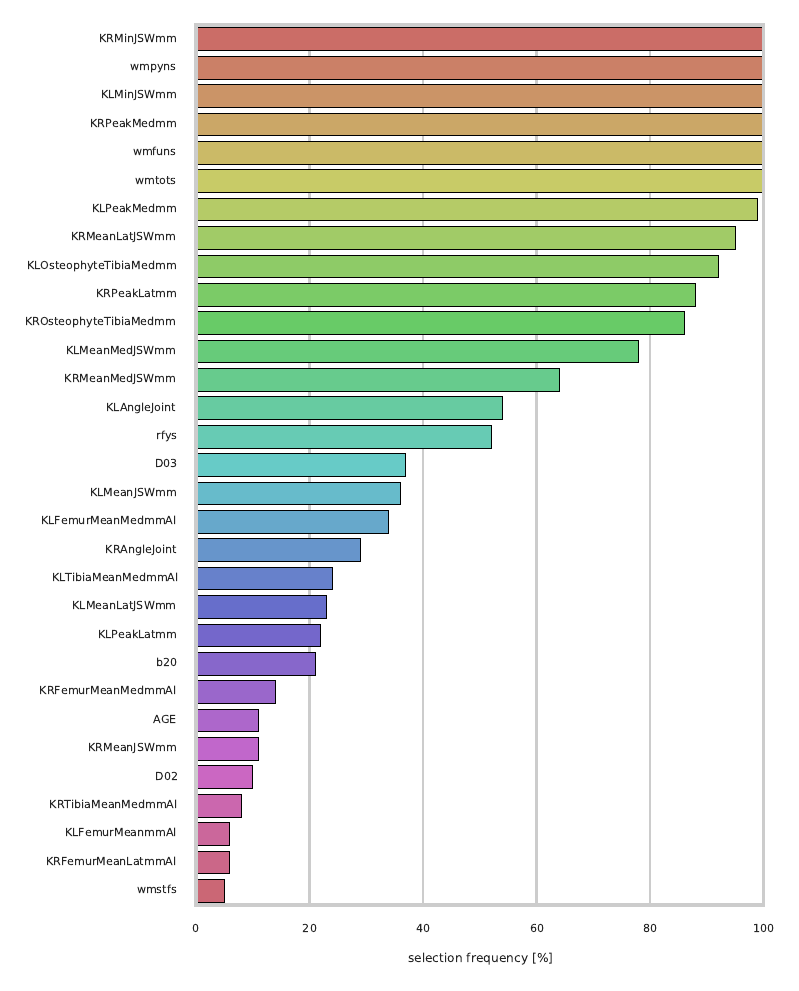}
	\caption{Frequency of feature selection with RFE procedure (\textbf{CHECK} dataset).}
	\label{fig:rfe-selected}
\end{figure}

\clearpage

\begin{figure}
	\centering
	\includegraphics[width=0.8\textwidth]{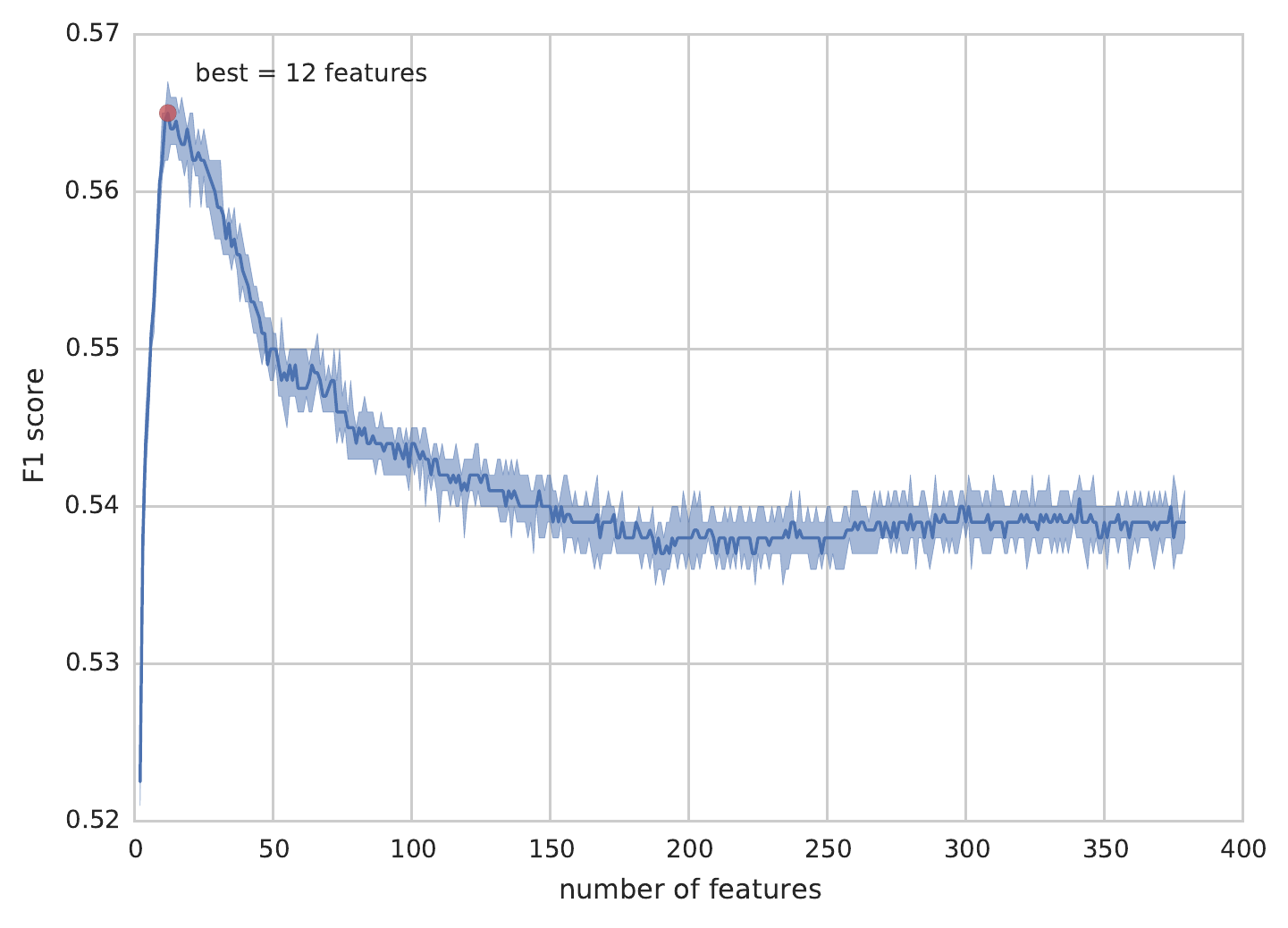}
	\caption{RFE scores (\textbf{CHECK} dataset).}
	\label{fig:rfe-check}
\end{figure}

\begin{figure}
	\centering
	\includegraphics[width=0.8\textwidth]{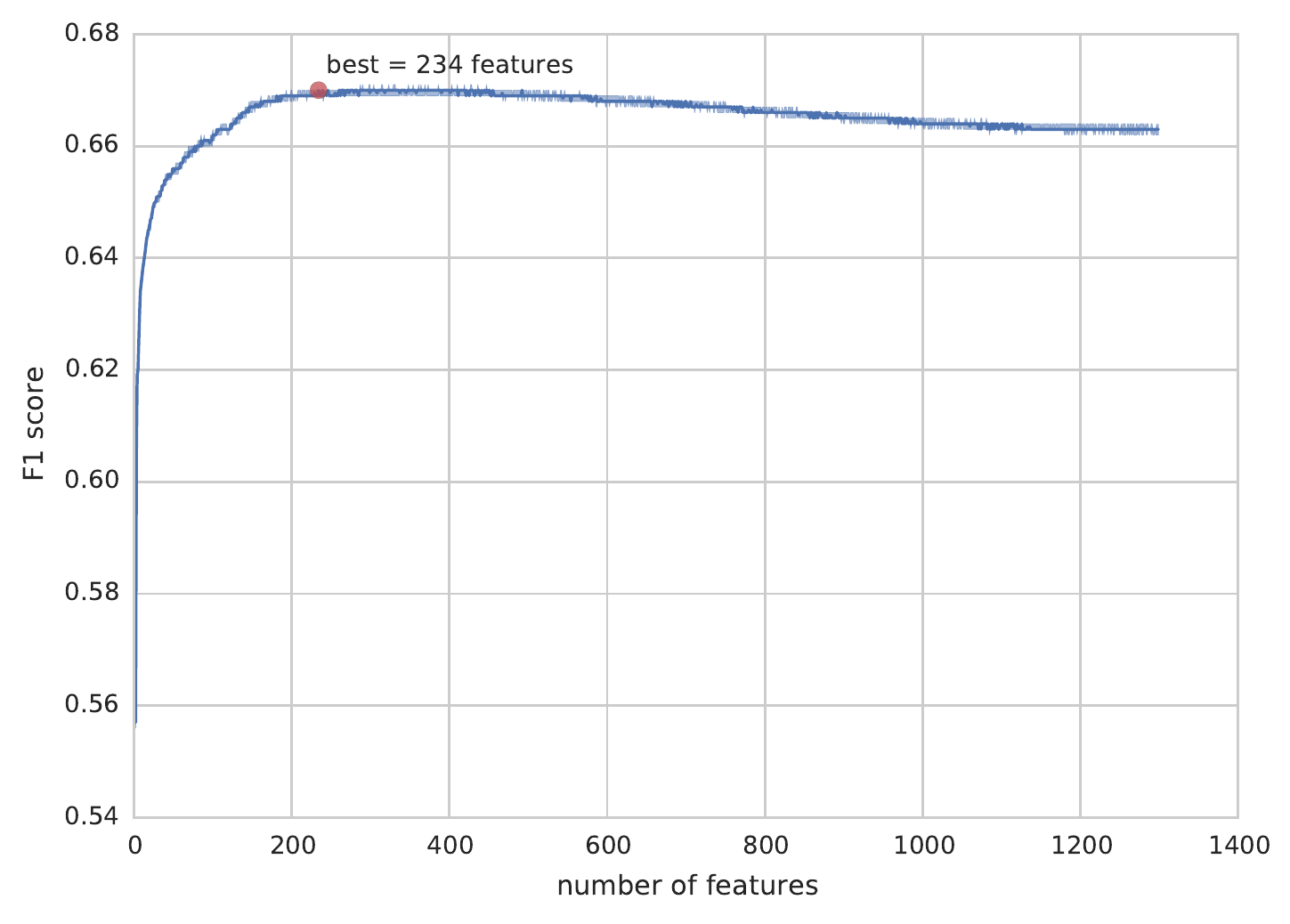}
	\caption{RFE scores (\textbf{OAI} dataset).}
	\label{fig:rfe-oai}
\end{figure}

\clearpage

\begin{figure}
	\centering
	\includegraphics[width=0.8\textwidth]{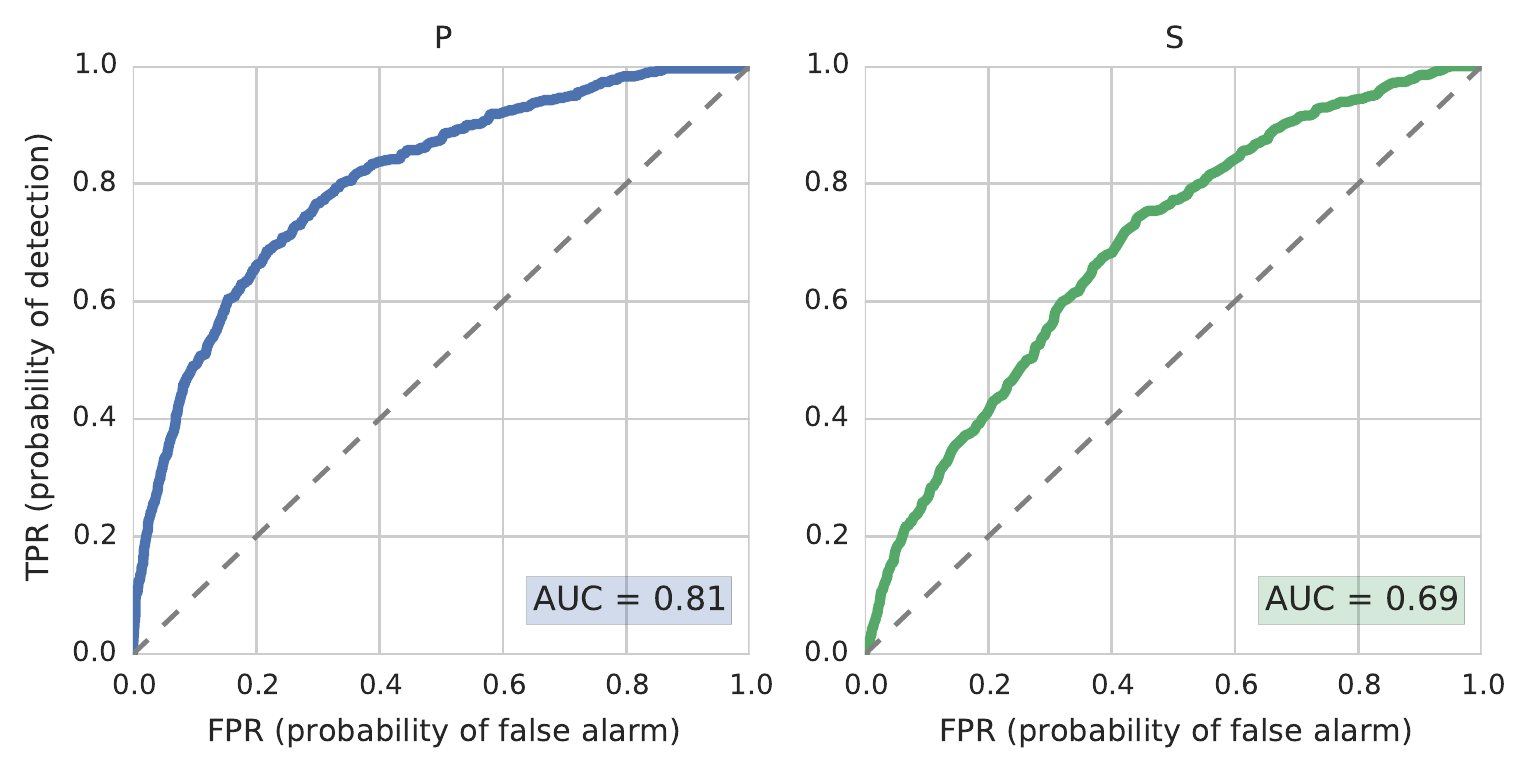}
	\caption{ROC curves for \textbf{P} and \textbf{S} sub-predictors of the best configuration of the \textit{duo classifier} (\textbf{CHECK} dataset).}
	\label{fig:roc-check}
\end{figure}

\begin{figure}
	\centering
	\includegraphics[width=0.8\textwidth]{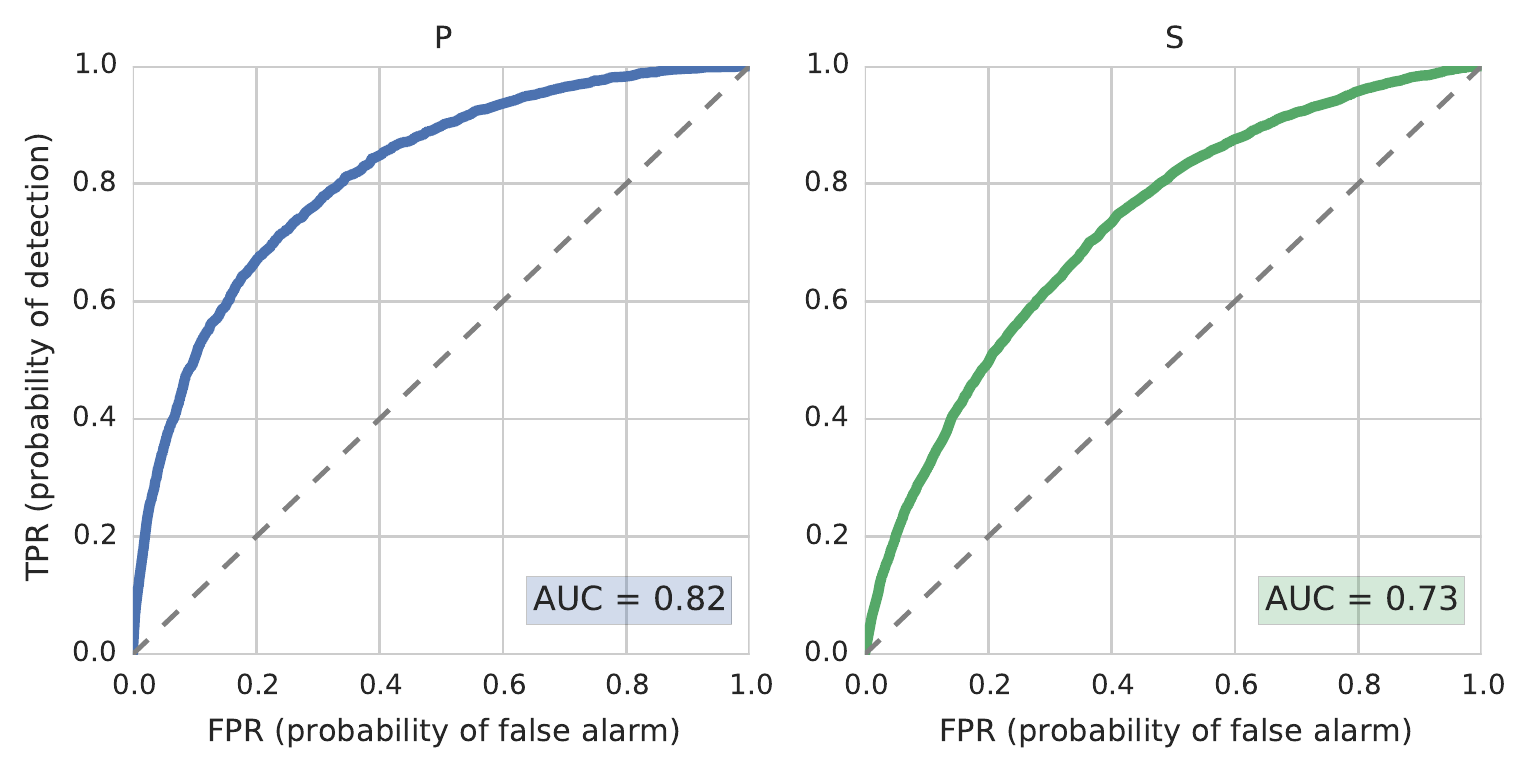}
	\caption{ROC curves for \textbf{P} and \textbf{S} sub-predictors of the best configuration of the \textit{duo classifier} (\textbf{OAI} dataset).}
	\label{fig:roc-oai}
\end{figure}

\clearpage

\begin{table}[!h]
\begin{subtable}{0.5\textwidth}
	\caption{\textbf{CHECK} dataset}
	\resizebox{0.963\columnwidth}{!}{
	\begin{tabular}{ll}
	\toprule
	AGE & age \\
	b20 & how good is your health today \\
	BEN & \textit{undocumented} (EQ-5D health survey) \\
	BMI & body mass index \\
	D02 & knee/hip pain intensity right know \\
	D03 & knee/hip pain intensity last week \\
	geen & not using medication \\
	HENDOL & left hip endorotation range of motion \\
	HPIJN & hip pain \\
	HSTIJF & morning stiffness in the hip \\
	I09a & do you feel limited in your role as a partner \\
	I09g & do you feel limited in fulfilling volunteering work \\
	KFLEXL & left knee flexion range of motion \\
	KLAngleJoint & left knee angle between the femur and tibia \\
	KLFemurMeanMedmmAl & left knee mean medial femur bone density \\
	KLMeanMedJSWmm & left knee mean medial joint space width \\
	KLMinJSWmm & left knee minimum total joint space width \\
	KLOsteophyteTibiaLatmm & left knee lateral tibia osteophyte area \\
	KLOsteophyteTibiaMedmm & left knee medial lateral tibia osteophyte area \\
	KLTibiaMeanMedmmAl & left knee mean medial tibia bone density \\
	KPIJN & knee pain \\
	KRAngleJoint & right knee angle between the femur and tibia \\
	KRMeanMedJSWmm & right knee mean medial joint space width \\
	KRMinJSWmm & right knee minimum total joint space width \\
	KROsteophyteTibiaMedmm & right knee medial lateral tibia osteophyte area \\
	KRPeakMedmm & right knee medial tibial eminence height \\
	KSTIJF & morning stiffness in the knee \\
	mobility & mobility (EQ-5D health survey) \\
	MVH\_A1 & index based on MVH-A1 value set (EQ-5D health survey) \\
	pain & pain or discomfort (EQ-5D health survey) \\
	PCIpiekr & worring (Pain Coping Inventory) \\
	rfys & physical functioning (SF-36 health survey) \\
	rment & general mental health (SF-36 health survey) \\
	rpijn & bodily pain (SF-36 health survey) \\
	rvit & vitality (SF-36 health survey) \\
	wmfuns & physical functioning sub-score (WOMAC) \\
	wmpyns & pain sub-score (WOMAC) \\
	wmstfs  & stiffness sub-score (WOMAC) \\
	wmtots & total score (WOMAC) \\
	\bottomrule
	\end{tabular}
	}
	\label{tab:features-check}
\end{subtable}
\begin{subtable}{0.5\textwidth}
	\caption{\textbf{OAI} dataset}
	\resizebox{\textwidth}{!}{
	\begin{tabular}{ll}
	\toprule
	DFBCOLL & difference in minutes between baseline and follow-up blood collection times \\
	DFUCOLL & difference in minutes between baseline and follow-up urine collection times \\
	DILKN16 & difficulty of heavy chores in last week (WOMAC) \\
	DIRKN12 & difficulty of lying down in teh last 7 days (WOMAC) \\
	DIRKN6 & pain level while walking in the last 7 days (WOMAC) \\
	GLCFQCV & glucosamine frequency of use in past 6 months \\
	HSPSS & physical summary score (SF-12 health survey) \\
	KGLRS & how much the knee pain and arthritis affect you? \\
	KIKBALL\_3.0 & leg used to kick a bal \\
	KOOSKPL & left knee pain score (KOOS) \\
	KOOSKPR & right knee pain score (KOOS) \\
	KOOSQOL & quality of life score (KOOS) \\
	KOOSYMR & symptoms score (KOOS) \\
	KPRKN2 & pain while fully straightening the knee in the last 7 days (KOOS) \\
	P7LKACV & average left knee pain in the last 7 days \\
	P7LKRCV & left knee pain severity in the last 7 days \\
	P7RKACV & right knee pain severity in the last 7 days \\
	P7RKRCV & average right knee pain in the last 7 days \\
	PMLKRCV & left knee pain severity in the last 30 days \\
	PMRKRCV & right knee pain severity in the last 30 days \\
	S1\_CFWDTH & width of femoral condyles used to define x = 1.0 \\
	S1\_IMPIXSZ & pixel size used for convertion to millimetres \\
	S1\_JSW150 & medial JSW at x = 0.15mm \\
	S1\_JSW175 & medial JSW at x = 0.175mm \\
	S1\_JSW200 & medial JSW at x = 0.2mm \\
	S1\_MCAJSW & average medial joint space width \\
	S1\_MCMJSW & minimum medial joint space width \\
	S1\_TMJSW & total minimum joint space width \\
	S1\_TPCFDS & distance from tibial plateau to tibial rim closest to femoral condyle \\
	S2\_IMPIXSZ & pixel size used for convertion to millimetres \\
	S2\_JSW150 & medial JSW at x = 0.15mm \\
	S2\_JSW175 & medial JSW at x = 0.175mm \\
	WOMADLL & left knee disability score (WOMAC) \\
	WOMADLR & right knee disability score (WOMAC) \\
	WOMKPL & left knee pain score (WOMAC) \\
	WOMKPR & right knee pain score (WOMAC) \\
	WOMTSL & left knee total score (WOMAC) \\
	WOMTSR & right knee total score (WOMAC) \\
	WPLKN3 & knee pain at night while in bed in the last 7 days \\
	WPLKN4 & knee pain sitting or lying down in the last 7 days \\
	\bottomrule
	\end{tabular}
	}
	\label{tab:features-oai}
\end{subtable}
\caption{Description of attributes shown in impact plots (\Cref{fig:shap-check,fig:shap-oai}).}
\end{table}

\end{document}

%% file: tables/best_rfe.tex
\begin{tabular}{lrlrr}
\toprule
 & & & \multicolumn{2}{l}{\textbf{$F_1$ score}} \\
\cmidrule(r){4-5}
\textbf{dataset} & \multicolumn{2}{l}{\textbf{features}} & \scriptsize{\textbf{median}} & \scriptsize{\textbf{95\% CI}} \\
\midrule
CHECK & 379 & (all) & 0.584 & (0.583, 0.586) \\
& 10--15 & (subset) & 0.573 & (0.570, 0.575) \\
\midrule
OAI & 1299 & (all) & 0.689 & (0.689, 0.690) \\
& 209--364 & (subset) & 0.676 & (0.675, 0.676) \\
\bottomrule
\end{tabular}

%% file: tables/expert-check.tex
\begin{tabular}{lrrrrrrrrrrrrr}
\toprule
 & \multicolumn{3}{r}{\textbf{N} \small{(1704)}} & \multicolumn{3}{r}{\textbf{P} \small{(358)}} & \multicolumn{3}{r}{\textbf{S} \small{(579)}} & \multicolumn{3}{r}{\textbf{P+S} \small{(160)}} & \textbf{not N} \\
\cmidrule(r){2-4} \cmidrule(r){5-7} \cmidrule(r){8-10} \cmidrule(r){11-13} \cmidrule(lr){14-14}
\textbf{selection} & \scriptsize{\textbf{abs}} & \scriptsize{\textbf{rel}} & \scriptsize{\textbf{recall}} & \scriptsize{\textbf{abs}} & \scriptsize{\textbf{rel}} & \scriptsize{\textbf{recall}} & \scriptsize{\textbf{abs}} & \scriptsize{\textbf{rel}} & \scriptsize{\textbf{recall}} & \scriptsize{\textbf{abs}} & \scriptsize{\textbf{rel}} & \scriptsize{\textbf{recall}} & \scriptsize{\textbf{recall}} \\
\midrule
conventional & 88 & 31\% & 5\% & 103 & 37\% & 29\% & 40 & 14\% & 7\% & 49 & 18\% & 31\% & 18\% \\
\texttt{ML-L} & 296 & 38\% & 17\% & 183 & 24\% & 51\% & 203 & 26\% & 35\% & 96 & 12\% & 60\% & 44\% \\

\bottomrule
\end{tabular}

%% file: tables/expert-oai.tex
\begin{tabular}{lrrrrrrrrrrrrr}
\toprule
 & \multicolumn{3}{r}{\textbf{N} \small{(12489)}} & \multicolumn{3}{r}{\textbf{P} \small{(951)}} & \multicolumn{3}{r}{\textbf{S} \small{(2718)}} & \multicolumn{3}{r}{\textbf{P+S} \small{(626)}} & \textbf{not N} \\
\cmidrule(r){2-4} \cmidrule(r){5-7} \cmidrule(r){8-10} \cmidrule(r){11-13} \cmidrule(lr){14-14}
\textbf{selection} & \scriptsize{\textbf{abs}} & \scriptsize{\textbf{rel}} & \scriptsize{\textbf{recall}} & \scriptsize{\textbf{abs}} & \scriptsize{\textbf{rel}} & \scriptsize{\textbf{recall}} & \scriptsize{\textbf{abs}} & \scriptsize{\textbf{rel}} & \scriptsize{\textbf{recall}} & \scriptsize{\textbf{abs}} & \scriptsize{\textbf{rel}} & \scriptsize{\textbf{recall}} & \scriptsize{\textbf{recall}} \\
\midrule
conventional & 858 & 52\% & 7\% & 366 & 22\% & 38\% & 187 & 11\% & 7\% & 229 & 14\% & 37\% & 18\% \\
\texttt{ML-L} & 2254 & 53\% & 18\% & 521 & 12\% & 55\% & 1059 & 25\% & 39\% & 385 & 9\% & 62\% & 46\% \\

\bottomrule
\end{tabular}

%% file: tables/expert-probabilities.tex
\begin{tabular}{llrrrr}
\toprule
\textbf{dataset} & \textbf{selection} & \textbf{N} & \textbf{P} & \textbf{S} & \textbf{P+S} \\
\midrule
CHECK & conventional & 31.4\% & 36.8\% & 14.3\% & 17.5\% \\
 & \texttt{ML-P} & 25.4\% & 28.2\% & 22.5\% & 23.6\% \\

\cmidrule(r){2-6}
OAI & conventional & 52.3\% & 22.3\% & 11.4\% & 14.0\% \\
 & \texttt{ML-P} & 38.5\% & 21.6\% & 22.3\% & 17.5\% \\

\bottomrule
\end{tabular}

%% file: tables/training-check.tex
\begin{tabular}{lllllll}
\toprule
 & \multicolumn{2}{l}{\textbf{33.3\% size}} & \multicolumn{2}{l}{\textbf{66.7\% size}} & \multicolumn{2}{l}{\textbf{100\% size}} \\
\cmidrule(r){2-3} \cmidrule(r){4-5} \cmidrule(r){6-7}
 & \scriptsize{\textbf{median}} & \scriptsize{\textbf{95\% CI}} & \scriptsize{\textbf{median}} & \scriptsize{\textbf{95\% CI}} & \scriptsize{\textbf{median}} & \scriptsize{\textbf{95\% CI}}\\
\midrule
\textbf{knn} & 0.325 & (0.308, 0.352) & 0.329 & (0.319, 0.350) & 0.338 & (0.327, 0.346) \\
\textbf{logreg} & 0.361 & (0.331, 0.370) & 0.35 & (0.337, 0.364) & 0.364 & (0.339, 0.377) \\
\textbf{logreg-multi} & 0.355 & (0.334, 0.367) & 0.377 & (0.348, 0.397) & 0.389 & (0.371, 0.418) \\
\textbf{random forest} & \textbf{0.399} & (0.378, 0.409) & \textbf{0.408} & (0.389, 0.427) & \textbf{0.425} & (0.411, 0.437) \\
\textbf{svc} & 0.341 & (0.306, 0.352) & 0.341 & (0.322, 0.362) & 0.366 & (0.336, 0.396) \\
\textbf{svc-rbf} & 0.36 & (0.318, 0.383) & 0.361 & (0.336, 0.379) & 0.365 & (0.350, 0.393) \\
\bottomrule
\end{tabular}

%% file: tables/training-oai.tex
\begin{tabular}{lllllll}
\toprule
 & \multicolumn{2}{l}{\textbf{33.3\% size}} & \multicolumn{2}{l}{\textbf{66.7\% size}} & \multicolumn{2}{l}{\textbf{100\% size}} \\
\cmidrule(r){2-3} \cmidrule(r){4-5} \cmidrule(r){6-7}
 & \scriptsize{\textbf{median}} & \scriptsize{\textbf{95\% CI}} & \scriptsize{\textbf{median}} & \scriptsize{\textbf{95\% CI}} & \scriptsize{\textbf{median}} & \scriptsize{\textbf{95\% CI}}\\
\midrule
\textbf{knn} & 0.369 & (0.359, 0.379) & 0.38 & (0.378, 0.387) & 0.389 & (0.379, 0.396) \\
\textbf{logreg} & 0.399 & (0.377, 0.402) & 0.415 & (0.400, 0.421) & 0.419 & (0.412, 0.425) \\
\textbf{logreg-multi} & 0.31 & (0.298, 0.317) & 0.323 & (0.310, 0.339) & 0.338 & (0.322, 0.343) \\
\textbf{random forest} & \textbf{0.417} & (0.408, 0.431) & \textbf{0.43} & (0.425, 0.437) & \textbf{0.437} & (0.435, 0.443) \\
\textbf{svc} & 0.36 & (0.341, 0.371) & 0.358 & (0.351, 0.362) & 0.375 & (0.363, 0.386) \\
\textbf{svc-rbf} & 0.41 & (0.387, 0.419) & 0.412 & (0.398, 0.425) & 0.426 & (0.418, 0.435) \\
\bottomrule
\end{tabular}

%% file: tables/training-multi-check.tex
\begin{tabular}{lllllll}
\toprule
 & \multicolumn{2}{l}{\textbf{37.5\% size}} & \multicolumn{2}{l}{\textbf{62.5\% size}} & \multicolumn{2}{l}{\textbf{100\% size}} \\
\cmidrule(r){2-3} \cmidrule(r){4-5} \cmidrule(r){6-7}
 & \scriptsize{\textbf{median}} & \scriptsize{\textbf{95\% CI}} & \scriptsize{\textbf{median}} & \scriptsize{\textbf{95\% CI}} & \scriptsize{\textbf{median}} & \scriptsize{\textbf{95\% CI}}\\
\midrule
\textbf{1vsR} & 0.491 & (0.489, 0.491) & 0.498 & (0.497, 0.499) & 0.502 & (0.501, 0.502) \\
\textbf{duo} & \textbf{0.494} & (0.493, 0.495) & \textbf{0.504} & (0.503, 0.505) & \textbf{0.507} & (0.506, 0.508) \\
\textbf{multilabel} & 0.489 & (0.488, 0.489) & 0.492 & (0.491, 0.493) & 0.496 & (0.495, 0.497) \\
\textbf{single} & 0.485 & (0.484, 0.486) & 0.49 & (0.489, 0.491) & 0.494 & (0.493, 0.495) \\
\bottomrule
\end{tabular}

%% file: tables/training-multi-oai.tex
\begin{tabular}{lllllll}
\toprule
 & \multicolumn{2}{l}{\textbf{42.9\% size}} & \multicolumn{2}{l}{\textbf{71.4\% size}} & \multicolumn{2}{l}{\textbf{100\% size}} \\
\cmidrule(r){2-3} \cmidrule(r){4-5} \cmidrule(r){6-7}
 & \scriptsize{\textbf{median}} & \scriptsize{\textbf{95\% CI}} & \scriptsize{\textbf{median}} & \scriptsize{\textbf{95\% CI}} & \scriptsize{\textbf{median}} & \scriptsize{\textbf{95\% CI}}\\
\midrule
\textbf{1vsR} & 0.64 & (0.640, 0.641) & 0.641 & (0.641, 0.641) & 0.642 & (0.642, 0.642) \\
\textbf{duo} & \textbf{0.644} & (0.643, 0.644) & \textbf{0.645} & (0.644, 0.645) & \textbf{0.646} & (0.646, 0.646) \\
\textbf{multilabel} & 0.639 & (0.638, 0.639) & 0.639 & (0.639, 0.639) & 0.641 & (0.641, 0.641) \\
\textbf{single} & 0.638 & (0.638, 0.639) & 0.639 & (0.639, 0.639) & 0.64 & (0.639, 0.640) \\
\bottomrule
\end{tabular}